\documentclass[journal]{IEEEtran}
\usepackage{amsmath}
\usepackage{amssymb}
\usepackage{amsmath,amsfonts}
\usepackage{algorithmic}
\usepackage{algorithm}
\usepackage{array}
\usepackage[caption=false,font=normalsize,labelfont=sf,textfont=sf]{subfig}
\usepackage{textcomp}
\usepackage{stfloats}
\usepackage{url}
\usepackage{verbatim}
\usepackage{graphicx}
\usepackage{cite}
\usepackage{mathtools}
\usepackage{xcolor}
\usepackage{tikz}
\usepackage{hyperref}
\usepackage{graphicx}     
\usepackage{booktabs}     
\usepackage{multirow}     
\usepackage{arydshln}

\usepackage{amsthm}  
\newtheorem{theorem}{Theorem}

\newtheorem{definition}{Definition}
\newtheorem{proposition}[theorem]{Proposition}

\theoremstyle{definition}
\newtheorem{assumption}[theorem]{Assumption}
\theoremstyle{remark}

\ifCLASSINFOpdf
\else
\fi

\begin{document}
  \title{Fine-Grained Instruction-Guided Graph Reasoning for Vision-and-Language Navigation}

\author{
Yaohua~Liu,
Binkai~Ou,
Xinyuan Song,
Rong~Fu, 
Amir H. Gandomi and  Simon Fong

\thanks{Y. Liu is with Chinese Medicine Guangdong Laboratory, Hengqin, Zhuhai 519031, China (Corresponding author: Yaohua Liu, e-mail: liuyaohua@hqcmlab.cn).}
\thanks{B. Ou is with the BoardWare Information System Company Ltd, Macau 999078, China (e-mail: yoanmike666@gmail.com).}
\thanks{X. Song  is with the Department of Computer Science, Emory University,
USA.}
\thanks{Rong Fu is with the Institute of Collaborative Innovation, University
    of Macau, Macau 999078, China (e-mail: mc46603@um.edu.mo).}
\thanks{Amir H. Gandomi is with the Faculty of Engineering \& IT, University
    of Technology Sydney, Ultimo, NSW 2007, Australia, and also
    with the University Research and Innovation Center (EKIK), Óbuda
    University, 1034 Budapest, Hungary (e-mail: gandomi@uts.edu.au).}
\thanks{Simon Fong is with the Faculty of Science and Technology, University
    of Macau, Macau 999078, China (e-mail: ccfong@um.edu.mo).}
}

  \maketitle

  \begin{abstract}    
Vision-and-Language Navigation (VLN) requires an embodied agent to traverse complex environments by following natural language instructions, demanding accurate alignment between visual observations and linguistic guidance. 
Despite recent progress, 
existing methods typically encode visual and directional cues in a coupled manner, and process instructions without explicitly extracting navigation-critical semantics, which often leads to imprecise spatial reasoning and suboptimal cross-modal alignment.
To address these challenges, we propose a fine-grained instruction-guided graph reasoning framework (FIGR) that enhances both spatial representation and instruction understanding during navigation. Specifically, an observation–graph interaction mechanism is introduced to disentangle angular and visual cues while strengthening directed edge representations through geometric embedding, enabling more reliable spatial reasoning within the navigation graph. The key-detail guidance module is implemented as Adaptive Open-Vocabulary Guidance (AOVG), where a contextual role parser dynamically identifies location, object, spatial-relation, and other contextual cues. This design avoids exact string matching and supports previously unseen entities and compositional expressions. For multilingual instructions, a Multilingual Semantic Adapter (MSA) maps language-specific representations into a shared navigation-semantic space. By jointly integrating structured graph reasoning with instruction-critical semantic cues, the proposed approach significantly improves the agent’s ability to follow complex navigation instructions. On the validation-unseen splits, FIGR achieves 67\% SPL on R2R and 64.8 sDTW on RxR, exceeding SPENav by 1 percentage point in SPL and PRET by 2.4 points in sDTW, respectively.
  \end{abstract}


  \IEEEpeerreviewmaketitle

  \section{Introduction}
  \IEEEPARstart{V}{ision}-and-Language Navigation (VLN)~\cite{gu2022vision,lin2025navcot,yu2025mossvln,chen2025constraint,wen2025ovl,liu2024volumetric,wu2024vision} is a challenging embodied intelligence task that requires an agent to navigate within complex three-dimensional environments~\cite{anderson2018vision,ku2020room} by executing natural language instructions. Successful navigation depends on the agent’s ability to jointly interpret visual observations of its surroundings and linguistic descriptions that specify spatial relations, landmarks, and target locations. This process involves reasoning over scene geometry, recognizing objects and structural cues, and making sequential decisions guided by the alignment between visual perception and instruction semantics. Unlike purely vision-based navigation, VLN demands robust cross-modal understanding, where subtle discrepancies between observation and language can accumulate over time and lead to navigation failure, particularly in long-horizon and instruction-dense scenarios~\cite{wang2023skill,qiao2023hop+,liu2023bird}. Consequently, effective integration of visual representations and fine-grained linguistic guidance is central to achieving accurate and reliable navigation behavior in VLN tasks.

  Recent advances~\cite{chen2022think,gao2023adaptive,hwang2023meta,wang2023gridmm,an2023bevbert,lu2024pret} in Vision-and-Language Navigation have increasingly leveraged explicit spatial representations to improve long-horizon decision making. In particular, a line of work introduces topological graph structures that record explored viewpoints and their connectivity, enabling agents to reason beyond purely local observations. Within such graph-based formulations, each node typically corresponds to a navigable viewpoint, while edges encode spatial transitions, allowing information to be propagated across the navigation space using graph neural networks or transformer-based encoders. This paradigm supports more strategic planning by aggregating visual context and linguistic guidance over previously visited locations. For example, DUET~\cite{chen2022think} combines global graph-level encoding with fine-grained local observation modeling to balance long-term planning and immediate action selection, while BEVBert~\cite{an2023bevbert} integrates local metric representations with global topological maps to enhance spatial awareness and cross-modal reasoning. More recently, PRET~\cite{lu2024pret} proposes a directed graph formulation that incorporates orientation-aware features into trajectory planning, offering an alternative to conventional undirected graph encoding and enabling more structured navigation decisions.

Despite the substantial progress achieved by graph-based VLN methods, several fundamental challenges remain unresolved. A primary limitation lies in the way visual observations~\cite{radford2021learning,oquab2024dinov2,li2023kerm,wang2023lana} and angular information are commonly represented: many existing approaches encode these heterogeneous cues within a single feature space, despite their distinct semantic roles in navigation. Such entangled representations may introduce interference between directional reasoning and visual perception, thereby reducing the precision of spatial decision making. In addition, navigation instructions~\cite{li2021align,conneau2019unsupervised,li2023improving,lin2023learning} are often treated as holistic textual inputs and integrated into the navigation process through generic attention mechanisms. This coarse-grained strategy fails to explicitly capture instruction-critical elements, such as location descriptors and object references, which are essential for resolving where the agent should move and what it should attend to at each step. As a result, current methods may struggle to maintain consistent instruction-following behavior, particularly in complex environments with long trajectories and densely specified instructions.

To address the aforementioned limitations, we propose a new framework for Vision-and-Language Navigation based on fine-grained instruction-guided graph reasoning (FIGR). The instruction branch combines Adaptive Open-Vocabulary Guidance (AOVG), which replaces exact string matching with contextual role parsing over location, object, spatial-relation, and other cues, with a Multilingual Semantic Adapter (MSA), which maps English, Hindi, and Telugu instruction features into a shared navigation-semantic space. By jointly strengthening spatial representation, open-vocabulary instruction grounding, and multilingual semantic consistency, the framework is designed to improve navigation decisions and trajectory alignment on VLN tasks. 

This article is an extended version of our prior conference paper~\cite{xie2025vln}. Compared with the conference version, this journal extension includes more detailed theoretical derivations and a deeper analysis of experimental results, providing a clearer understanding of the proposed method and its empirical behavior across different navigation settings. In addition, the fixed location/object vocabulary used by the conference key-detail guidance is replaced by AOVG, and the RxR evaluation is extended with an MSA-based multilingual formulation for English, Hindi, and Telugu instructions.

The main contributions of this work can be summarized as follows:
\begin{enumerate}
    \item We propose FIGR, a fine-grained instruction-guided graph reasoning framework that jointly strengthens spatial reasoning and instruction-level semantic alignment. Its observation–graph interaction module decouples angular and visual cues and incorporates geometric information into directed edge representations for more reliable candidate-node reasoning.
    \item We develop an instruction-guidance branch that combines Adaptive Open-Vocabulary Guidance (AOVG) and a Multilingual Semantic Adapter (MSA). AOVG replaces fixed-vocabulary matching with contextual parsing of location, object, spatial-relation, and other cues, while MSA maps English, Hindi, and Telugu instructions into a shared navigation-semantic space.
    \item We provide a unified theoretical analysis linking feature decoupling, contextual semantic grounding, and multilingual adaptation to reduced representational interference, improved optimization stability, and stronger instruction–trajectory alignment.
\end{enumerate}

The rest of this paper is organized as follows. Section II reviews related work on Vision-and-Language Navigation and graph-based spatial representations. Section III describes the proposed fine-grained instruction-guided graph reasoning framework in detail. Section IV presents the theoretical analysis of the proposed method. Section V reports experimental results and ablation studies. Finally, Section VI concludes the paper and discusses future research directions. 

  \section{Related Works}
  \subsection{Vision-and-Language Navigation}
VLN~\cite{fried2018speaker,wang2019reinforced,chen2021history,chen2022think,gao2023adaptive,hwang2023meta,wang2023gridmm,an2023bevbert,lu2024pret} has attracted increasing attention in recent years as a core problem in embodied artificial intelligence, where agents are required to execute natural language instructions in visually rich environments. Early VLN approaches predominantly relied on sequence-to-sequence architectures based on recurrent neural networks, mapping textual instructions to either low-level navigation actions or high-level waypoint decisions using discretized panoramic visual observations. While effective in simple scenarios, these methods often suffered from limited generalization and insufficient long-horizon reasoning capability.

To alleviate data scarcity and improve robustness, subsequent studies introduced various data augmentation techniques including back-translation~\cite{fried2018speaker} and environment-level dropout~\cite{tan2019learning,li2022envedit,liu2021vision}. Attention-based mechanisms~\cite{vaswani2017attention,xie2023cross,xie2024pointtalk} were also explored to enhance cross-modal alignment during navigation.

With the emergence of transformer architectures, recent VLN research has shifted toward more expressive representation learning and planning strategies. PREVALENT~\cite{hao2020towards} pioneers large-scale pretraining for VLN by initializing transformer encoders on navigation-related data, significantly improving generalization to unseen environments. MTVM~\cite{lin2022multimodal} further extends this direction by incorporating navigation history memorization within a transformer framework, allowing agents to reason over variable-length trajectories and develop more robust planning behaviors.

Beyond representation learning, navigation strategy design has become a central focus, particularly for improving exploration and decision making in unseen environments. DUET~\cite{chen2022think} introduces a dual-scale transformer architecture that jointly models local observations and global topological context, enabling balanced short-term action selection and long-term planning. AZHP~\cite{gao2023adaptive} proposes a hierarchical graph structure to improve exploration efficiency, while Meta-Explore~\cite{hwang2023meta} explicitly models backtracking decisions to enhance trajectory recovery. BEVBert~\cite{an2023bevbert} integrates local metric maps with global topological graphs, strengthening spatial understanding through hybrid mapping representations. More recently, PRET~\cite{lu2024pret} formulates VLN planning using directed graph structures with orientation-aware trajectory encoding, offering an alternative to conventional graph convolution-based map representations and enabling more structured navigation reasoning. 
In contrast to these methods, our work builds upon directed graph formulations and addresses two underexplored aspects: (1) decoupling angular and visual cues within graph representations, and (2) explicitly extracting navigation-critical semantics from instructions for tighter cross-modal alignment.
Our framework considers the multilingual nature of RxR. Rather than maintaining independent semantic vocabularies for different languages, we use a shared-role formulation with lightweight language-conditioned adapters. This design separates language-specific surface variation from the language-independent spatial and object semantics needed for graph-based navigation.

\subsection{Maps for Navigation}
Spatial mapping has long been a central component of visual navigation systems, with early research primarily grounded in simultaneous localization and mapping (SLAM) techniques~\cite{fuentes2015visual,wang2024survey}. 
Classical approaches construct metric maps encoding geometric layouts~\cite{chaplot2020object,henriques2018mapnet,xie2024hecpg}, enabling precise localization but often incurring substantial computational overhead that limits scalability.

To improve efficiency, subsequent studies explored topological map representations~\cite{chaplot2020neural,chen2021topological}, where environments are modeled as graphs composed of discrete nodes and navigable connections. Compared to dense metric maps, topological graphs offer a more compact and computationally efficient abstraction for global planning, albeit at the cost of reduced local geometric precision. This trade-off motivated the development of hybrid mapping strategies that combine metric and topological representations, aiming to exploit detailed local geometry while retaining efficient long-range reasoning.

Recent advances~\cite{wang2021structured,an2023bevbert,chen2022think} have further introduced learnable mapping frameworks that integrate spatial representations into end-to-end navigation models. For example, several works incorporate neural metric maps for short-term spatial reasoning while maintaining global topological graphs to support long-horizon planning. 
BEVBert~\cite{an2023bevbert} exemplifies this direction by jointly learning local metric and global topological representations.
However, processing hybrid maps often remains computationally demanding, particularly when dense metric representations are repeatedly updated during navigation.

To address these challenges, alternative formulations~\cite{chen2021topological,lu2024pret} have investigated graph-centric navigation strategies that avoid explicit metric map construction. Directed topological maps and trajectory-based planning methods enable incremental embedding updates and reduce computational complexity while preserving effective spatial reasoning capabilities. By encoding orientation-aware transitions and trajectory fidelity within directed graphs, these approaches support structured planning without relying on dense geometric maps. Building on this line of work, our method adopts a directed graph formulation and further enhances edge representations through observation–graph interaction, enabling more accurate and efficient spatial reasoning for vision-and-language navigation.

\begin{figure*}[!t]
	\centerline{\includegraphics[width=1.0\linewidth]{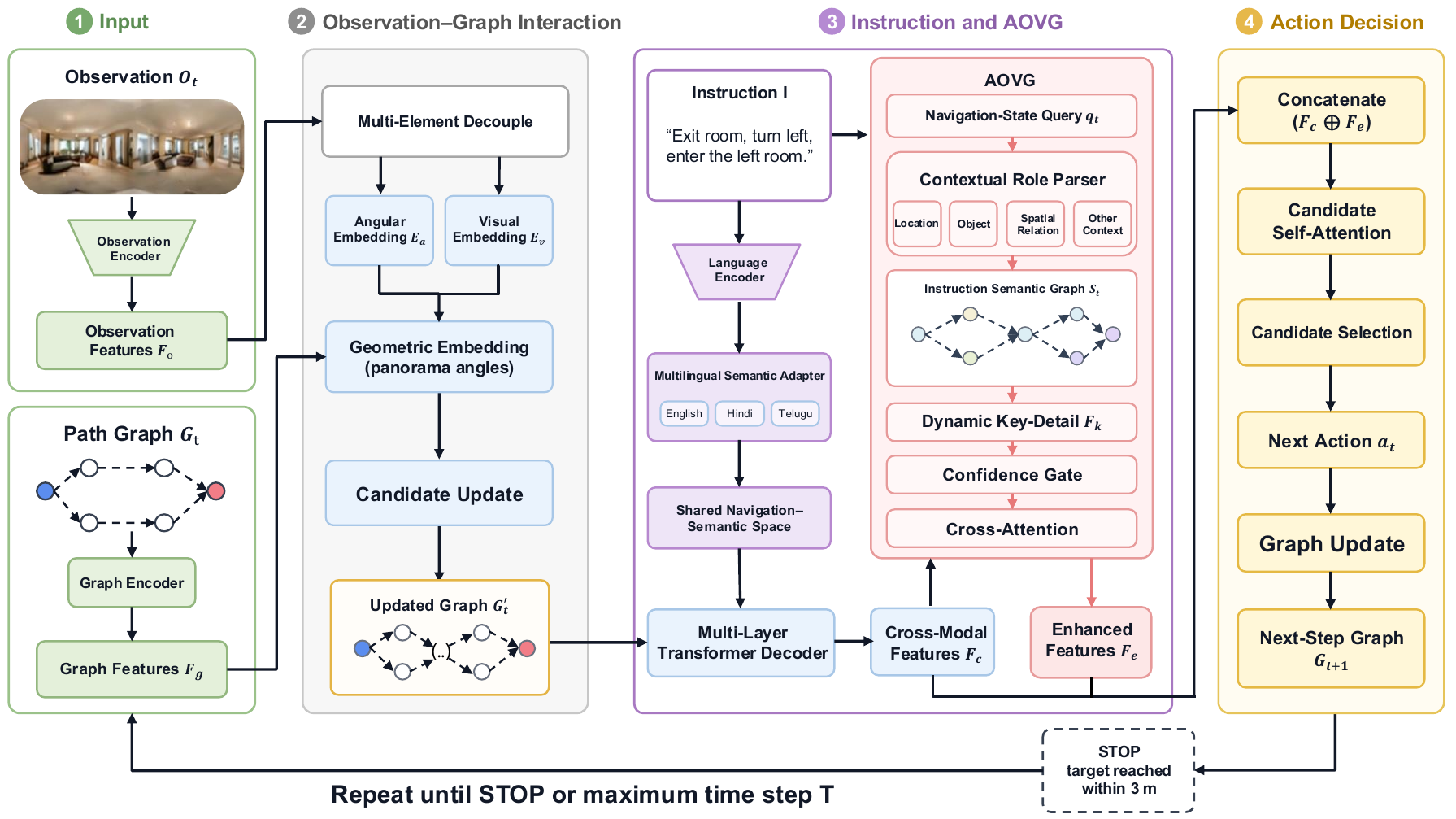}}
	\vspace{-0.1cm}
	\caption{Illustration of the proposed FIGR architecture. At time step $t$, given the observation $O_t$ and path graph $G_t$, observation features $F_o$ and graph features $F_g$ are extracted and processed by the Observation--Graph Interaction module, which strengthens directed-edge representations and updates candidate nodes. The resulting graph is fused with instruction features. The Adaptive Open-Vocabulary Guidance module (AOVG) dynamically parses location, object, spatial-relation, and other contextual cues, while the Multilingual Semantic Adapter maps English, Hindi, and Telugu representations into a shared semantic space. These instruction-aware features guide candidate selection, after which the selected action updates the path graph to $G_{t+1}$.}
	\label{fig:pipeline}
	\vspace{-0.1cm}
\end{figure*}

 \section{METHOD}

\subsection{Problem Statement}
VLN~\cite{anderson2018vision,ku2020room,chen2022think} is formulated in a discrete environment~\cite{chang2017matterport3d} represented by a navigation graph ${G}=\{\mathcal{V},\mathcal{E}\}$, where $\mathcal{V} = \{V_i\}_{i=1}^N$ denotes the set of navigable nodes and $\mathcal{E}$ represents the traversable connections between nodes. An agent is initialized at a starting node and is provided with a natural language instruction $I = \{I_i\}_{i=1}^M$ consisting of $M$ words, which specifies the target destination through spatial and semantic descriptions. The objective of the agent is to follow the instruction and navigate through the graph to reach the target location.

At each time step $t$, the agent located at node $V_t$ perceives its surroundings through a 360-degree panoramic observation  ${O}_t = \{r_{t,i}\}_{i=1}^K$, where each $r_{t,i}$ corresponds to the visual feature extracted from the $i$-th viewpoint. The agent also has access to a set of navigable neighboring nodes $\mathcal{N}(V_t)\subset \mathcal{V}$, along with their relative spatial coordinates and orientations with respect to the current position. Based on the observed visual information, spatial geometry, and linguistic guidance, the agent must decide whether to transition to one of the neighboring nodes or remain at its current location.

Navigation is considered successful when the agent terminates within a predefined Euclidean distance threshold (3 meters) from the target position. The task therefore requires the agent to balance perception, language understanding, and sequential decision making over a finite horizon $T$, while maintaining consistent cross-modal alignment between visual observations and instruction semantics throughout the navigation process.

\subsection{Model Overview}
Fig.~\ref{fig:pipeline} presents an overview of the proposed fine-grained instruction-guided graph reasoning framework. At each time step $t$, the agent receives a panoramic visual observation $O_t$, from which observation features $F_o$ are extracted using a pretrained visual encoder~\cite{oquab2024dinov2}. In parallel, a navigation graph ${G_t}=\{\mathcal{V}_t,\mathcal{E}_t\}$ is maintained to represent the agent’s explored environment, where nodes correspond to navigable viewpoints and directed edges encode orientation-aware transitions. Unlike prior approaches~\cite{chen2022think,an2023bevbert} that directly assign panoramic features to visited nodes and directional views to unvisited nodes, leading to inconsistent representations, we adopt a directed graph formulation that incrementally incorporates nodes and edges as navigation proceeds. This design ensures a ified and consistent representation for both visited and candidate nodes. 

The navigation graph features $F_g$ are constructed from directed edges that capture both visual observations and angular information associated with specific movement directions. Since observation and graph features inherently contain heterogeneous cues, including visual appearance and orientation, directly fusing them may introduce representational interference. To address this issue, we introduce an observation–graph interaction mechanism (Sec.~\ref{sec:ogi}) that explicitly decouples angular and visual information within observation features. The angular cues derived from panoramic observations are further integrated into the graph representation through geometric embedding, strengthening the expressiveness of directed edges and enabling more accurate spatial reasoning. Based on this interaction, candidate node features are iteratively updated, resulting in an enhanced graph representation $G'_t$.

Given the updated graph $G'_t$, the navigation instruction $I$ is incorporated to guide action selection. Specifically, the graph features are fused with instruction features through a multi-layer transformer decoder to obtain cross-modal representations that encode the alignment between linguistic semantics and navigable trajectories. The key-detail guidance module is implemented as AOVG, which dynamically parses open-vocabulary spans into location, object, spatial-relation, and other contextual roles. For multilingual instructions, a MSA maps language-specific features into a shared navigation-semantic space before cross-modal enhancement. These cues are used to enhance cross-modal representations via a dedicated interaction mechanism, producing instruction-aware features that emphasize decision-relevant information.

Finally, a candidate selection module predicts the next navigation action by scoring candidate nodes based on their alignment with the enhanced instruction-aware features. The selected node is used to update the navigation graph $G_{t+1}$, and the process is repeated until the agent terminates or reaches the maximum time step. The complete inference procedure is summarized in Algorithm~\ref{alg:figr}. Through the integration of observation–graph interaction and fine-grained instruction guidance, the proposed framework enables consistent spatial reasoning and precise instruction following throughout the navigation process.

\begin{algorithm}[t]
\caption{Inference procedure of the proposed FIGR framework.}
\label{alg:figr}
\begin{algorithmic}[1]
\REQUIRE Instruction $I$, initial node $V_1$, initial graph $G_1$, maximum step $T$
\ENSURE Predicted trajectory $\tau$
\STATE Encode $I$ with the language encoder; for RxR, compute MSA features $\widetilde H^{\ell}$.
\STATE Initialize $\tau\leftarrow[V_1]$ and $t\leftarrow1$.
\WHILE{$t\le T$ and the termination condition is false}
    \STATE Observe the panorama $O_t$ and extract visual features $F_o$.
    \STATE Construct graph features $F_g$ from the candidate nodes and directed edges of $G_t$.
    \STATE Decouple angular and visual cues in $F_o$ and compute geometric embeddings for $F_g$.
    \STATE Apply the candidate-query/observation-key-value decoder to obtain the updated graph $G'_t$.
    \STATE Fuse $G'_t$ with the instruction features to obtain cross-modal features $F_c^{(t)}$.
    \STATE Use the state query $q_t$ to parse contextual spans into the four AOVG roles.
    \STATE Build the semantic relation graph $\mathcal S_t$ and compute $F_k^{(t)}$.
    \STATE Compute the confidence gate $g_t$ and enhanced features $F_e^{(t)}$.
    \STATE Score candidate nodes using $F_e^{(t)}$ and select the next node $V_{t+1}$.
    \STATE Update $G_{t+1}$ with $V_{t+1}$ and append $V_{t+1}$ to $\tau$.
    \STATE $t\leftarrow t+1$.
\ENDWHILE
\RETURN $\tau$
\end{algorithmic}
\end{algorithm}

\subsection{Observation-Graph Interaction}\label{sec:ogi}
Given the observation features $F_o$ and graph features $F_g$, we introduce an observation–graph interaction mechanism to enhance directed edge representations and iteratively update candidate node features. The observation features extracted from panoramic views inherently contain heterogeneous information, including visual appearance and angular cues. Treating these components as a unified representation may lead to mutual interference, as angular information primarily supports directional reasoning, whereas visual features encode semantic and appearance-related cues. To mitigate this issue, we explicitly decouple the observation features into angular and visual components and process them separately.

Specifically, the angular and visual components are independently embedded to obtain angular embeddings $E_a$ and visual embeddings $E_v$, respectively. These embeddings are then concatenated and fused through a multilayer perceptron to form the refined observation representation:
\begin{equation}
{F'_o} =  \mathcal{F}(\mathrm{cat}[E_a,E_v]),
\end{equation}
where $\mathrm{cat} [\cdot]$ denotes the concatenation and  $\mathcal{F}(\cdot)$ represents the fusion function implemented by an MLP. This decoupling-and-fusion strategy reduces representational interference while preserving complementary information from both modalities.

For graph features, we focus on angular information to model directional perception, as orientation plays a dominant role in navigation decisions. The angular representation consists of both heading and elevation angles, where heading information is emphasized due to its greater influence on horizontal navigation. For each candidate node, we extract the sine and cosine representations of the heading angle $\alpha$, denoted as $\mathrm{sin}(\alpha)$ and $\mathrm{cos}(\alpha)$. Correspondingly, we obtain the angular representations $\mathrm{sin}(\alpha ')$ and $\mathrm{cos}(\alpha ')$ from the observation features. The relative angular difference is computed using a trigonometric formulation:
\begin{equation}
d =  |(\mathrm{atan2}[\mathrm{sin}(\alpha-\alpha'),\mathrm{cos}(\alpha-\alpha')])|.
\end{equation}.

Based on the relative angular differences, we identify the closest angular distance $d'$ for each candidate view and construct a geometric positional embedding:
\begin{equation}\label{eq:3}
pe =  \mathcal{P}(\mathrm{cat}[d', \mathrm{sin}(\alpha-\alpha'), \mathrm{cos}(\alpha-\alpha']),
\end{equation}
where $\mathcal{P}(\cdot)$ denotes a linear projection function. This positional embedding encodes the spatial relationship between candidate nodes and the current observation, enabling the model to distinguish candidate directions more effectively.

The final candidate node features are obtained by combining the raw angular information with the positional embedding through element-wise addition:
\begin{equation}\label{eq:4}
F'_g =  \mathcal{C}(\mathrm{cat}[\mathrm{sin}(\alpha), \mathrm{cos}(\alpha), \mathrm{sin}(\beta), \mathrm{cos}(\beta)]) + pe,
\end{equation}
where $\beta$ denotes the elevation angle and $\mathcal{C}(\cdot)$ is a linear transformation. This formulation enriches the graph representation with explicit geometric priors while maintaining a compact feature structure.

To update directed edge features and incorporate observation-level context, we employ a transformer decoder layer~\cite{vaswani2017attention} for observation–graph interaction. In this process, candidate node features serve as queries, while the refined observation features $F'_o$ act as keys and values. This attention-based interaction allows each directed edge to selectively attend to observation regions corresponding to its spatial orientation, resulting in an updated navigation graph $G'_t$ with strengthened edge representations and improved directional consistency.
\begin{figure}[t]
\centerline{\includegraphics[width=1.0\linewidth]{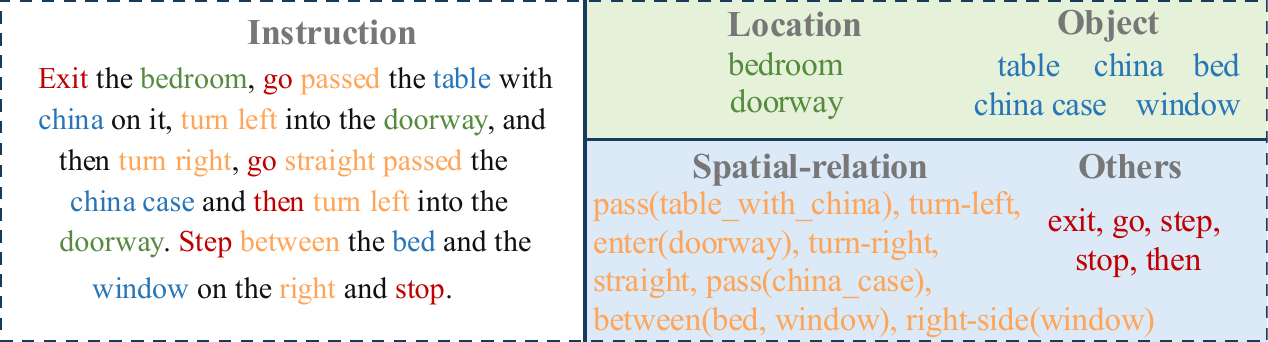}}
\vspace{-0.1cm}
\caption{The process of dynamically parsing location, object, spatial-relation, and other contextual roles from the instruction in the AOVG module.}
\vspace{-0.3cm}
\label{fig:lo_ob}
\end{figure}

\subsection{Key-Detail Guidance}\label{sec:kdg}
After obtaining the updated navigation graph $G'_t$, the natural language instruction $I$ is incorporated to guide the agent’s navigation decisions. We first fuse the graph representation with instruction features using a multi-layer transformer decoder, producing cross-modal features $F_c^{(t)}$ that encode interactions between navigable trajectories and linguistic semantics. To overcome the vocabulary-coverage limitation of exact string matching, we design an Adaptive Open-Vocabulary Guidance (AOVG) module. AOVG parses instruction spans in the context of the current navigation state, rather than assigning semantics solely according to membership in a pre-defined word list.

Let $F_i=\{h_1,\ldots,h_M\}$ denote the instruction features and let $q_t$ be a query representing the current navigation state:
\begin{equation}
q_t=\operatorname{MLP}\!\left(\left[\operatorname{Pool}(F_o^{\prime(t)}),\operatorname{Pool}(F_g^{\prime(t)}),\operatorname{Pool}(F_c^{(t)})\right]\right).
\end{equation}
As illustrated in Fig. \ref{fig:lo_ob}, for each candidate span $s_{ij}=\operatorname{SpanEnc}(h_i,\ldots,h_j)$, a contextual role parser predicts four semantic roles, $\mathcal R=\{\mathrm{location},\mathrm{object},\mathrm{spatial\mbox{-}relation},\mathrm{other}\}$:
\begin{equation}
p(r_{ij}\mid s_{ij},q_t)=\operatorname{Softmax}\!\left(W_r[s_{ij};q_t]+b_r\right),\qquad r_{ij}\in\mathcal R.
\end{equation}
The spatial-relation role includes directional, metric, and ordinal subtypes (e.g., left/right, front/behind, near/far, before/after, first/second/last). Consequently, open-vocabulary expressions such as ``alcove'', ``balcony'', and ``the second door on the left'' can be represented as contextual spans and compositional relations even when their surface forms are absent from an offline vocabulary.

\begin{equation}
Z_t^{r}=\sum_{i\le j}a_{t,ij}\,p(r_{ij}=r\mid s_{ij},q_t)\,s_{ij},\qquad r\in\mathcal R,
\end{equation}
where $a_{t,ij}$ is a state-conditioned span-attention weight. The extracted spans additionally form an instruction semantic graph $\mathcal S_t=(\mathcal X_t,\mathcal E_t^I)$, whose nodes $\mathcal X_t$ are location/object spans and whose edges $\mathcal E_t^I$ encode spatial and ordinal relations.
\begin{equation}
p(u_{ab}\mid x_a,x_b,q_t)=\operatorname{Softmax}\!\left(W_u[x_a;x_b;q_t]+b_u\right),
\end{equation}
where $u_{ab}$ denotes a relation subtype. This graph representation preserves the compositional structure of expressions such as ``the second door on the left'', instead of treating ``second'' and ``left'' as unrelated tokens.

The dynamic key-detail representation is obtained by fusing the four role embeddings and the semantic-graph encoding:
\begin{equation}
\begin{gathered}F_k^{(t)}=\mathcal F_{\mathrm{AOVG}}\!\big([Z_t^{\mathrm{location}},Z_t^{\mathrm{object}},Z_t^{\mathrm{spatial}},\\[-1mm] Z_t^{\mathrm{other}},\operatorname{GNN}(\mathcal S_t)]\big).\end{gathered}
\end{equation}
Unlike the original fixed location/object vocabulary, AOVG can retain an unmatched or uncertain status when an entity is not visible or cannot be grounded reliably. A confidence gate therefore controls the contribution of dynamic guidance:
\begin{equation}
\begin{gathered}g_t=\sigma\!\left(W_g[q_t;F_k^{(t)}]+b_g\right),\\[-1mm] F_e^{(t)}=F_c^{(t)}+g_t\odot\operatorname{CrossAttn}(F_c^{(t)},F_k^{(t)}).\end{gathered}
\end{equation}
The enhanced features are further refined by self-attention across candidate nodes, and an MLP computes the semantic compatibility of each candidate with the instruction. The selected action updates $G_{t+1}$. AOVG is trained with the navigation objective and auxiliary role, relation, and grounding losses; no external language model or fixed vocabulary is required at inference time.

\subsection{Multilingual Semantic Adapter}\label{sec:msa}
Natural-language navigation instructions are expressed in many languages. Even when different languages describe the same trajectory, lexical and syntactic variation can induce language-dependent semantic shifts. Using the English, Hindi, and Telugu instructions in RxR as a representative multilingual setting, we introduce a Multilingual Semantic Adapter (MSA) after the language encoder and before AOVG. Let $\ell\in\{\mathrm{en},\mathrm{hi},\mathrm{te}\}$ denote the language and let $H_i^{\ell}\in\mathbb R^{M\times D_t}$ be the corresponding token features. Each language has a lightweight residual adapter, while the output projection and semantic-role heads are shared:
\begin{equation}
\begin{gathered}A_{\ell}(H_i^{\ell})=H_i^{\ell}+U_{\ell}\,\sigma(V_{\ell}H_i^{\ell}),\\[-1mm] U_{\ell}\in\mathbb R^{D_t\times d_a},\quad V_{\ell}\in\mathbb R^{d_a\times D_t}.\end{gathered}
\end{equation}
\begin{equation}
\begin{gathered}\widetilde H_i^{\ell}=\operatorname{LN}\!\left(W_sA_{\ell}(H_i^{\ell})+W_{\ell}e_{\ell}+b_s\right),\\[-1mm] \ell\in\{\mathrm{en},\mathrm{hi},\mathrm{te}\}.\end{gathered}
\end{equation}
\begin{equation}
\begin{gathered}
\widetilde H_i^{\mathrm{en}}=\operatorname{LN}\!\left(W_sA_{\mathrm{en}}(H_i^{\mathrm{en}})+W_{\mathrm{en}}e_{\mathrm{en}}+b_s\right),\\
\widetilde H_i^{\mathrm{hi}}=\operatorname{LN}\!\left(W_sA_{\mathrm{hi}}(H_i^{\mathrm{hi}})+W_{\mathrm{hi}}e_{\mathrm{hi}}+b_s\right),\\
\widetilde H_i^{\mathrm{te}}=\operatorname{LN}\!\left(W_sA_{\mathrm{te}}(H_i^{\mathrm{te}})+W_{\mathrm{te}}e_{\mathrm{te}}+b_s\right).
\end{gathered}
\end{equation}
where $e_{\ell}$ is a learned language embedding, $W_s$ is shared across languages, and $W_{\ell}$ is a language-specific bias projection. The adapted features $\widetilde H_i^{\ell}$ replace $H_i$ in the AOVG parser, so the same role and relation heads are used for all three languages:
\begin{equation}
p(r_{ij}\mid s_{ij}^{\ell},q_t,\ell)=\operatorname{Softmax}\!\left(W_r[\widetilde s_{ij}^{\ell};q_t]+b_r\right),\qquad r\in\mathcal R.
\end{equation}
To encourage language-invariant grounding, we align the pooled semantic representations of language realizations describing the same RxR trajectory. For a minibatch containing languages $\ell$ and $\ell'$ for the same episode, we use:
\begin{equation}
\mathcal L_{\mathrm{msa}}=\left\|\operatorname{Pool}(\widetilde H_i^{\ell})-\operatorname{Pool}(\widetilde H_i^{\ell'})\right\|_2^2.
\end{equation}
The MSA does not translate instructions or require language-specific vocabularies. It learns small language-conditioned residual transformations while preserving a shared semantic space for AOVG roles, spatial relations, and graph-node alignment.

\subsection{Loss Function}
Following common training paradigms~\cite{hao2020towards,an2023bevbert,chen2022think,lu2024pret} in Vision-and-Language Navigation, we adopt a two-stage training strategy consisting of pretraining and task-specific fine-tuning. During the pretraining stage, Masked Language Modeling (MLM) is employed to initialize the model parameters. Specifically, 15\% of the input tokens are randomly masked and fed into the text encoder. The encoded representations, together with the corresponding directed edge features, are processed by a transformer decoder to reconstruct the masked tokens. This pretraining objective encourages the model to learn robust cross-modal representations that effectively ground linguistic semantics in the navigation graph.

After pretraining, the model is fine-tuned using a hybrid learning strategy that combines teacher forcing and student forcing. In the teacher-forcing phase, the agent is guided by the ground-truth action sequence $a_t^{gt}$, while still conditioning on the model’s predicted actions $a_t^p$ to maintain consistency between training and inference. In the student-forcing phase, pseudo actions $a'_t$ are sampled from the model’s predicted action distribution to promote exploration and alleviate exposure bias.

For trajectories that deviate from the ground-truth path, we adopt a recovery strategy by assigning pseudo-labels based on the nearest unvisited ground-truth node. When no such nodes are available, pseudo-labels are determined using the shortest path between the current position and the target location. This mechanism enables the model to learn corrective behaviors when deviations occur.

The overall training objective is defined as a weighted combination of losses from teacher forcing and student forcing:
\begin{equation}
\mathcal{L}_{\mathrm{nav}} = \lambda(\frac{1}{T} \sum_{t=1}^{T} \eta(a_t^p, a_t^{gt})) + (1-\lambda)(\frac{1}{T} \sum_{t=1}^{T} \eta(a_t^p, a'_t)),
\end{equation}
where $\lambda$ controls the balance between the two learning strategies, $T$ denotes the total number of time steps, and $\eta(\cdot)$ represents the cross-entropy loss. The navigation loss is supplemented with contextual-role, relation-grounding, and multilingual alignment objectives:
\begin{equation}
\mathcal{L}=\mathcal{L}_{\mathrm{nav}}+\lambda_r\mathcal{L}_{\mathrm{role}}+\lambda_s\mathcal{L}_{\mathrm{rel}}+\lambda_g\mathcal{L}_{\mathrm{ground}}+\lambda_m\mathcal{L}_{\mathrm{msa}}.
\end{equation}
Here, $\mathcal{L}_{\mathrm{role}}$ and $\mathcal{L}_{\mathrm{rel}}$ supervise the four semantic roles and relation subtypes, $\mathcal{L}_{\mathrm{ground}}$ encourages consistency between extracted spans and candidate graph nodes, and $\mathcal{L}_{\mathrm{msa}}$ is the language-invariant alignment loss defined in Sec.~\ref{sec:msa}. When role or relation annotations are unavailable, they are generated as training-time weak labels from instruction--trajectory alignment; no external language model is used during inference.

\section{Theoretical Analysis}

To further understand why the proposed FIGR framework improves navigation stability and decision accuracy, we present a unified analysis of three operations that are explicitly defined in the Method section: (i) angular--visual decoupling and geometric embedding in the Observation--Graph Interaction module, (ii) state-conditioned open-vocabulary grounding in AOVG, and (iii) language-conditioned projection into a shared semantic space by MSA. The analysis does not assume that a fixed location/object vocabulary is complete. Instead, it treats location, object, spatial-relation, and other spans as contextual random variables and accounts for the confidence gate used to suppress uncertain grounding. For RxR, all role variables are computed from the MSA-adapted features $\widetilde H_i^{\ell}$. Theorem~\ref{thm:ogi}, Theorem~\ref{thm:aovg}, and Proposition~\ref{prop:msa} therefore describe complementary effects: reduced spatial interference, improved instruction--trajectory alignment, and lower cross-lingual representation discrepancy.

\begin{definition}[Observation Feature Decoupling]\label{def:decouple}
Let \(F_o \in \mathbb{R}^{K \times D}\) be the observation feature matrix with \(K\) viewpoints and feature dimension \(D\). Assume
\begin{equation}
\begin{aligned}
F_o = [A, V], &\quad A \in \mathbb{R}^{K \times D_a}, \\
V \in \mathbb{R}^{K \times D_v}, &\quad D_a + D_v = D.
\end{aligned}
\end{equation}
Define the angular and visual embedding maps
\begin{equation}
E_a = \phi_a(A), \quad E_v = \phi_v(V),
\end{equation}
where \(\phi_a: \mathbb{R}^{D_a} \to \mathbb{R}^{D_e}\) and \(\phi_v: \mathbb{R}^{D_v} \to \mathbb{R}^{D_e}\) are fully connected layers with ReLU activation. The fused observation embedding is
\begin{equation}
F'_o = \mathcal{F}([E_a, E_v]) = \mathrm{MLP}([E_a, E_v]).
\end{equation}
\end{definition}

\begin{definition}[Angular Distance Embedding]\label{def:angle}
Let \(F_g \in \mathbb{R}^{N \times D_g}\) denote graph features with node angles \(\alpha \in [0,2\pi)^N\). Define
\begin{equation}
d_{ij} = \min_{k \in \mathbb{Z}} |\alpha_i - \alpha'_j + 2k\pi|.
\end{equation}
The angular distance \(d_{ij}\) represents the minimum circular angle displacement between viewpoints \(i\) and \(j\).
\end{definition}

\begin{definition}[Geometric Positional Embedding]\label{def:geom}
With \(d_{ij}\) defined in Definition~\ref{def:angle} and based on Equation~\eqref{eq:3}, define the geometric embedding
\begin{equation}
\psi(\alpha_i, \alpha'_j) = \mathcal{P}\big(d_{ij}, \sin(\alpha_i - \alpha'_j), \cos(\alpha_i - \alpha'_j)\big),
\end{equation}
and positional embedding
\begin{equation}
pe_{ij} = \psi(\alpha_i, \alpha'_j).
\end{equation}
The updated graph embedding in Equation~\eqref{eq:4} is
\begin{equation}\label{eq:13}
F'_g = \mathcal{C}\big(\sin(\alpha), \cos(\alpha), \sin(\beta), \cos(\beta)\big) + pe,
\end{equation}
where $\beta$ denotes the elevation angle. Thus, the theoretical representation includes the same heading--elevation terms as the graph update in Sec.~\ref{sec:ogi}, rather than modeling heading alone.
\end{definition}

\begin{assumption}[OGI regularity conditions]\label{assump:ogi}
The downstream navigation loss is twice differentiable and has a bounded Hessian in a neighborhood of the training representations. The angular and visual channels have bounded cross-covariance after $\phi_a$ and $\phi_v$, and the geometric map $\mathcal P$ is Lipschitz and preserves local circular angular differences. The candidate-query/observation-key-value decoder is also Lipschitz in its inputs.
\end{assumption}

\begin{theorem}[Effect of Decoupling and Geometric Embedding]\label{thm:ogi}
Let \(F'_o\) and \(F'_g\) be as defined in Definition~\ref{def:decouple}, Definition~\ref{def:angle}, and Definition~\ref{def:geom}, and let the updated candidate features be obtained by the transformer decoder described in Sec.~\ref{sec:ogi}. Under Assumption~\ref{assump:ogi}, if the reduction in angular--visual cross-covariance and the preserved geometric signal is nonzero, the expected gradient second moment of the navigation objective $\mathcal L_{\mathrm{nav}}$ satisfies
\begin{equation}
\mathbb{E}(\|\nabla_{F'_o, F'_g} \mathcal{L}_{\mathrm{nav}}\|^2)
\leq
\mathbb{E}(\|\nabla_{F_o, F_g} \mathcal{L}_{\mathrm{nav}}\|^2) - \delta,
\end{equation}
for some constant \(\delta > 0\), where the expectation is taken over the training data distribution. The constant $\delta$ depends on the removed cross-channel interference, the strength of the geometric signal, and the Lipschitz bounds; the statement is therefore conditional rather than an unconditional guarantee for every parameterization.
\end{theorem}

Theorem~\ref{thm:ogi} establishes that, under the stated regularity conditions, angular--visual decoupling and the heading--elevation geometric embedding reduce the expected gradient second moment relative to the directly coupled representation. The result explains why the candidate-query/observation-key-value interaction can learn directed-edge relationships with a better-conditioned spatial signal; it does not claim that the representation alone guarantees a particular success rate.

This theoretical result is consistent with the empirical evaluation on R2R and RxR, where FIGR improves success-related metrics and trajectory alignment scores. A complete derivation of the bound, including the cross-covariance and Lipschitz terms, is provided in Supplementary Material I.

Building on the spatial result, we next analyze the instruction branch. The following result covers the complete AOVG computation: state-conditioned span roles, relation-graph encoding, confidence gating, and cross-modal enhancement.

\begin{definition}[Contextual AOVG representation]\label{def:aovg}
Let $\bar H_i^{\ell}$ denote the instruction features entering AOVG, with $\bar H_i^{\ell}=H_i$ for monolingual inputs and $\bar H_i^{\ell}=\widetilde H_i^{\ell}$ after MSA. At time $t$, let $q_t$ be the navigation-state query and let
\[
\mathcal R=\{\mathrm{location},\mathrm{object},\mathrm{spatial\mbox{-}relation},\mathrm{other}\}.
\]
For each span $s_{ij}^{\ell}=\operatorname{SpanEnc}(\bar h_i^{\ell},\ldots,\bar h_j^{\ell})$, define
\begin{equation}
 p(r_{ij}\mid s_{ij}^{\ell},q_t,\ell)=\operatorname{Softmax}\!\left(W_r[s_{ij}^{\ell};q_t]+b_r\right),\qquad r_{ij}\in\mathcal R,
\end{equation}
and the role embeddings
\begin{equation}
 Z_t^r=\sum_{i\le j}a_{t,ij}\,p(r_{ij}=r\mid s_{ij}^{\ell},q_t,\ell)\,s_{ij}^{\ell},\qquad r\in\mathcal R.
\end{equation}
The span graph $\mathcal S_t=(\mathcal X_t,\mathcal E_t^I)$ encodes relation subtypes between extracted entities, and the AOVG key-detail representation and gated enhancement are
\begin{equation}
\begin{gathered}
F_k^{(t)}=\mathcal F_{\mathrm{AOVG}}\!\left([\{Z_t^r\}_{r\in\mathcal R},\operatorname{GNN}(\mathcal S_t)]\right),\\[-1mm]
g_t=\sigma\!\left(W_g[q_t;F_k^{(t)}]+b_g\right),\\[-1mm]
F_e^{(t)}=F_c^{(t)}+g_t\odot\operatorname{CrossAttn}(F_c^{(t)},F_k^{(t)}).
\end{gathered}
\end{equation}
\end{definition}

\begin{assumption}[AOVG grounding conditions]\label{assump:aovg}
The score $S(\cdot,G_{\mathrm{gt}})$ is locally Lipschitz in the enhanced representation, the cross-attention operator is bounded, and each correctly grounded role provides non-redundant information about the ground-truth path conditioned on $q_t$. Let $\epsilon_t$ upper-bound the contribution of unmatched or uncertain spans after confidence gating.
\end{assumption}

\begin{theorem}[Contextual Semantic Grounding and Alignment]\label{thm:aovg}
Let $F_c^{(t)}$, $F_k^{(t)}$, $g_t$, and $F_e^{(t)}$ be defined in Definition~\ref{def:aovg}. Under Assumption~\ref{assump:aovg}, there exist constants $\rho>0$ and $\epsilon_t\ge 0$ such that the AOVG enhancement satisfies
\begin{equation}
S(F_e^{(t)}, G_{\mathrm{gt}}) - S(F_c^{(t)}, G_{\mathrm{gt}})
\ge
\rho\sum_{r\in\mathcal R}\operatorname{MI}\!\left(Z_t^r;G_{\mathrm{gt}}\mid q_t\right)-\epsilon_t,
\end{equation}
where $\operatorname{MI}(\cdot;\cdot\mid\cdot)$ denotes conditional mutual information. The bound includes all four roles and the state query; it reduces to the location/object-only formulation only when the spatial-relation and other terms carry no additional information and $\epsilon_t=0$.
\end{theorem}

Theorem~\ref{thm:aovg} formalizes why contextual parsing is more expressive than exact string matching: compositional entities such as ``the second door on the left'' can contribute through both entity and relation variables, while unsupported or visually ungrounded spans are attenuated rather than forced into a candidate node. A complete derivation of the alignment bound is provided in Supplementary Material II.

\begin{proposition}[MSA reduces language-dependent discrepancy]\label{prop:msa}
For language-aligned RxR episodes describing the same trajectory, define
\begin{equation}
\Delta_{\mathrm{msa}}=\mathbb E_{\ell\ne\ell'}\left[\left\|\operatorname{Pool}(\widetilde H_i^{\ell})-\operatorname{Pool}(\widetilde H_i^{\ell'})\right\|_2^2\right]
\end{equation}
and let $\Delta_{\mathrm{id}}$ be the same discrepancy before adaptation. Assume that the MSA parameter class contains a no-adaptation (identity or shared-isometry) setting that reproduces $\Delta_{\mathrm{id}}$. Since $\mathcal L_{\mathrm{msa}}$ directly optimizes this discrepancy, any global minimizer over that class satisfies
\begin{equation}
\Delta_{\mathrm{msa}}\leq \Delta_{\mathrm{id}}.
\end{equation}
If the aligned episodes contain a nonzero language-dependent component that is representable by the adapters, the inequality is strict. This proposition concerns representation consistency; it does not by itself imply a guaranteed improvement in SR, nDTW, or sDTW.
\end{proposition}

Together, Theorem~\ref{thm:ogi}, Theorem~\ref{thm:aovg}, and Proposition~\ref{prop:msa} connect the three method components without conflating their roles: OGI stabilizes directional graph representations, AOVG injects state-conditioned semantic evidence with explicit uncertainty control, and MSA makes that evidence comparable across the English, Hindi, and Telugu settings of RxR.
\begin{table*}[!t]
	\renewcommand\arraystretch{1.0}
	\caption{Evaluation results on R2R dataset under different settings. Best performance is highlighted in bold.}
	\begin{center}
	    \resizebox{1.0\linewidth}{!}{
			\begin{tabular}{lcccccccccccc}
				\toprule
                & \multicolumn{4}{c}{\textit{Val Seen}} & \multicolumn{4}{c}{\textit{Val Unseen}} & \multicolumn{4}{c}{\textit{Test Unseen}}\\
			    {Methods}&TL&NE$\downarrow$&SR$\uparrow$&SPL$\uparrow$ &TL&NE$\downarrow$&SR$\uparrow$&SPL$\uparrow$ &TL&NE$\downarrow$&SR$\uparrow$&SPL$\uparrow$\\
				\midrule
				{Seq2Seq-SF~\cite{anderson2018vision}}  &11.33  &6.01 &39 &-   &8.39&7.81&22&-&8.13&7.85&28&18\\
                {Speaker-Follower~\cite{fried2018speaker}}    &-  &3.36  &66  &- &- &6.62&35&-&14.82&6.62&35&28\\
                {RCM~\cite{wang2019reinforced}}    &10.65  &3.53  &67  &- &11.46 &6.09&43&-&11.97&6.12&43&38\\
                {Regretful~\cite{ma2019regretful}}   &-  &3.23  &69  &63 &- &5.32&50&41&-&5.69&56&40\\
			{EnvDrop~\cite{tan2019learning}}    &11.00  &3.99  &62  &59 &10.70 &5.22&52&48&11.66&5.23&51&47\\
                {PREVALENT~\cite{hao2020towards}}   &10.32  &3.67  &69  &65 &10.19 &4.71&58&53&10.51&5.30&54&51\\
                {NvEM~\cite{an2021neighbor}}    &11.09  &3.44  &69  &65 &11.83 &4.27&60&55&12.98&4.37&58&54\\
                {SSM~\cite{wang2021structured}}    &14.70  &3.10  &71  &62 &20.70 &4.32&62&45&20.40&4.57&61&46\\
                {RecBert~\cite{hong2021vln}}    &11.13&2.90&72&68&12.01&3.93&63&57&12.35&4.09&63&57\\
                {HAMT~\cite{chen2021history}}    &11.15&2.51&76&72&11.46&2.29&66&61&12.27&3.93&65&60\\
                {MTVM~\cite{lin2022multimodal}}    &-&2.67&74&69&-&3.73&66&59&-&3.85&65&59\\
                {DUET~\cite{chen2022think}}    &12.32&2.28&79&73&13.94&3.31&72&60&14.73&3.65&69&59\\
                {AZHP~\cite{gao2023adaptive}}    &-&-&-&-&14.05&3.15&72&61&14.95&3.52&71&60\\
                {Meta-Explore~\cite{hwang2023meta}}    &11.95&2.11&81&75&13.09&3.22&72&62&14.25&3.57&71&61\\
                {GridMM~\cite{wang2023gridmm}}    &-&-&-&-&13.27&2.83&75&64&14.43&3.35&73&62\\
                {BEVBert~\cite{an2023bevbert}}    &13.56&2.17&81&74&14.55&2.81&75&64&15.87&3.13&73&62\\
                {PRET~\cite{lu2024pret}}    &11.25&2.41&78&72&11.87&2.90&74&65&12.21&3.09&72&64\\
                {SPENav~\cite{yuan2026spenav}}    &-&-&-&-&-&2.84&\textbf{76}&66&-&2.92&\textbf{76}&65\\
				\midrule
                {FIGR}    &11.12&\textbf{2.05}&\textbf{83}&\textbf{77}&11.68&\textbf{2.61}&\textbf{76}&\textbf{67}&12.73&\textbf{2.89}&75&\textbf{66}\\
				\bottomrule
			\end{tabular}}
	        \label{tab:r2r}
	\end{center}
\end{table*}

\section{EXPERIMENTS}

\subsection{Implementation Details}
We implement the proposed method using PyTorch~\cite{paszke2019pytorch} and conduct all experiments on a single NVIDIA A6000 GPU. Model optimization is performed using the AdamW~\cite{loshchilov2017decoupled} optimizer. During the pretraining stage, the learning rate is set to $2\text{e-}5$, the batch size is 16, and training is carried out for 100,000 iterations. For task-specific fine-tuning, the learning rate is reduced to $1\text{e-}5$, the batch size is set to 8, and the model is trained for 200,000 iterations. The weighting coefficient $\lambda$ in the loss function is fixed to 0.2 across all experiments.

For both the R2R and RxR datasets, we adopt augmented training data provided by prior works to improve generalization performance. Unless otherwise specified, all reported results are obtained using the same network architecture and training protocol, ensuring fair comparisons with existing methods~\cite{hao2020towards,wang2022less}.
AOVG is trained with weak role and relation labels obtained from instruction--trajectory alignment, while MSA uses language-aligned RxR episodes to construct $\mathcal{L}_{\mathrm{msa}}$. The language adapters are lightweight residual blocks and share the role and relation prediction heads. At inference time, the parser operates only on the encoded instruction and current navigation state; neither the offline location/object vocabulary nor an external large language model is required.

\subsection{Comparison on R2R Dataset}
\subsubsection{R2R Dataset}
The Room-to-Room (R2R)~\cite{anderson2018vision} dataset consists of 7,189 navigation trajectories collected from 90 indoor scenes, with each trajectory paired with three natural language instructions. The environments are represented by panoramic viewpoints, where each viewpoint comprises 36 discretized visual observations. The dataset is divided into training, validation seen, validation unseen, and test unseen splits. The validation seen split shares the same environments as the training set, while the validation unseen and test unseen splits evaluate generalization to novel scenes. All trajectories in R2R correspond to shortest paths between the start and target locations.

\subsubsection{Evaluation Metrics}
Following standard evaluation protocols~\cite{an2023bevbert,lu2024pret}, we report four metrics to assess navigation performance. (a) Trajectory Length (\textit{TL}) measures the average length of the executed navigation paths. (b) Navigation Error (\textit{NE}) denotes the mean shortest-path distance between the agent’s final position and the target location. (c) Success Rate (\textit{SR}) represents the percentage of episodes in which the agent terminates within 3 meters of the target. (d) Success weighted by Path Length (\textit{SPL}) further accounts for navigation efficiency by balancing success rate against trajectory length, with higher values indicating more effective navigation.

\subsubsection{Quantitative Results}
We compare the proposed method with a wide range of representative VLN approaches, including Seq2Seq-SF~\cite{anderson2018vision}, Speaker-Follower~\cite{fried2018speaker}, RCM~\cite{wang2019reinforced}, Regretful~\cite{ma2019regretful}, EnvDrop~\cite{tan2019learning}, PREVALENT~\cite{hao2020towards}, NvEM~\cite{an2021neighbor}, SSM~\cite{wang2021structured}, RecBert~\cite{hong2021vln}, HAMT~\cite{chen2021history}, MTVM~\cite{lin2022multimodal}, DUET~\cite{chen2022think}, AZHP~\cite{gao2023adaptive}, Meta-Explore~\cite{hwang2023meta}, GridMM~\cite{wang2023gridmm}, BEVBert~\cite{an2023bevbert}, PRET~\cite{lu2024pret} and SPENav~\cite{yuan2026spenav}. Quantitative results on the R2R benchmark are summarized in Table~\ref{tab:r2r}. On validation unseen, FIGR achieved an SR of 76\% and an SPL of 67\%, matching SPENav's SR while improving SPL by 1 percentage point. Its NE decreased from SPENav's 2.84\,m to 2.61\,m. On test unseen, FIGR achieved an SPL of 66\% and an NE of 2.89\,m, compared with SPENav's 65\% and 2.92\,m, respectively. However, its test-unseen SR was 75\%, 1 percentage point below SPENav, indicating an advantage in success-weighted path efficiency rather than uniformly higher success rates. The proposed method achieves competitive or superior performance across different evaluation splits, with particularly strong results on the SPL metric, which reflects both navigation accuracy and efficiency. These results indicate that fine-grained instruction-guided graph reasoning improves the agent’s ability to follow navigation instructions while maintaining efficient trajectories.

\begin{figure*}[!t]
	\centerline{\includegraphics[width=1.0\linewidth]{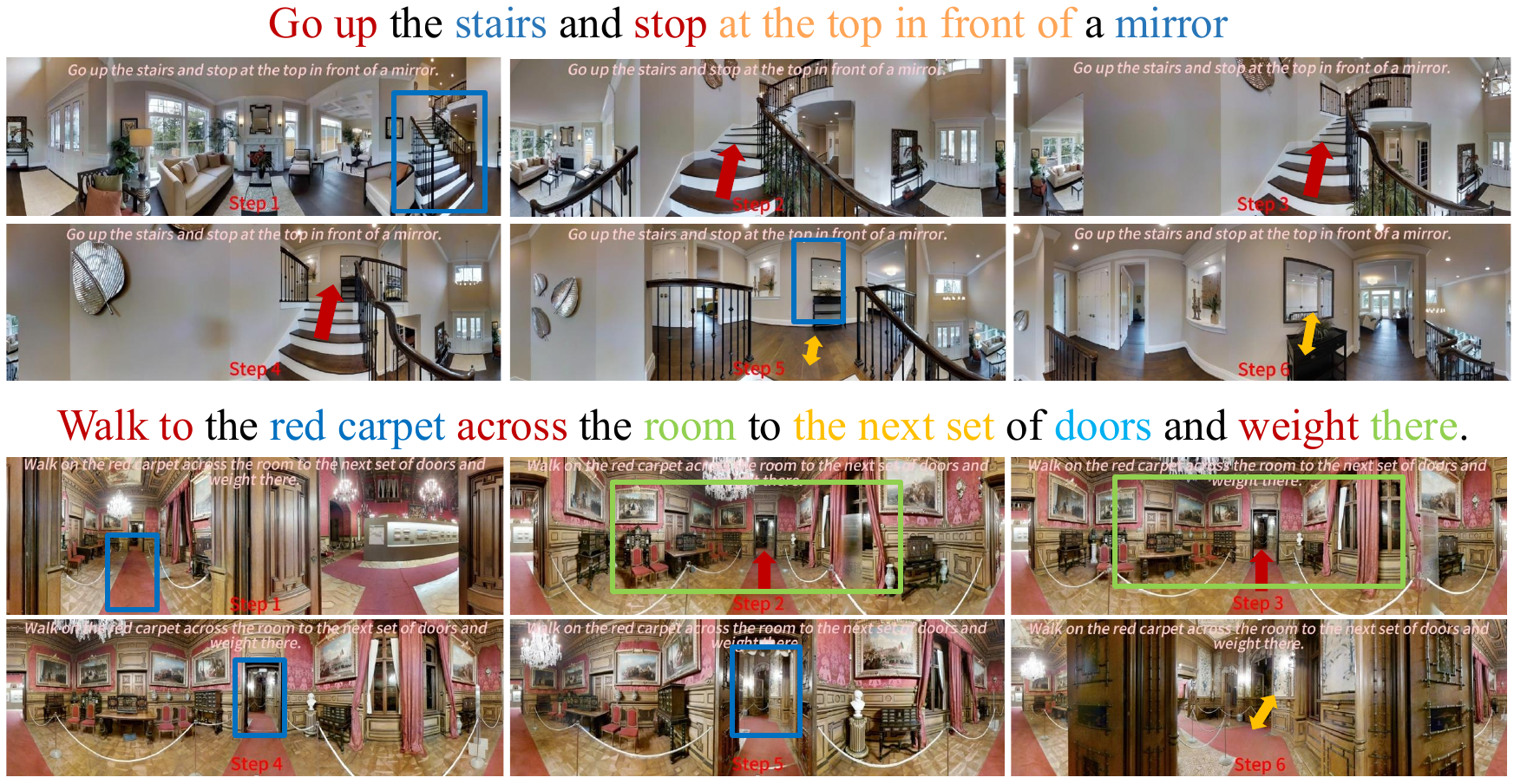}}
	\caption{Qualitative results on the R2R Dataset. The green box indicates the location details, and the blue box represents the object details.
    }
	\label{fig:demo}
\end{figure*}

In addition to quantitative evaluation, qualitative examples are presented in Fig.~\ref{fig:demo}, illustrating that the proposed method is able to accurately interpret instructions and generate coherent navigation trajectories in complex indoor environments.

\subsection{Comparison on RxR Dataset}
\subsubsection{RxR Dataset}
The Room-across-Room (RxR)~\cite{ku2020room} dataset is a large-scale multilingual benchmark for VLN. It contains a total of 126,000 navigation instructions evenly distributed across three languages: English, Hindi, and Telugu. Compared with R2R, navigation trajectories in RxR are substantially longer and do not necessarily correspond to shortest paths. Moreover, RxR instructions are more descriptive and linguistically complex, posing additional challenges for long-horizon reasoning and fine-grained instruction understanding. To evaluate multilingual adaptation, we retain the original RxR splits and report English-, Hindi-, and Telugu-specific results in addition to aggregate scores. The language-specific adapters are trained jointly with shared AOVG role and relation heads, allowing the effect of language adaptation to be separated from the effect of open-vocabulary parsing.
\begin{table}[!t]
	\renewcommand\arraystretch{1.0}
        \setlength{\tabcolsep}{2pt}
	\caption{Evaluation results on RxR dataset under different settings. Best performance is highlighted in bold.}
	\begin{center}
	    \resizebox{1.0\linewidth}{!}{
			\begin{tabular}{lcccccccc}
				\toprule
                & \multicolumn{4}{c}{\textit{Val Seen}} & \multicolumn{4}{c}{\textit{Val Unseen}}\\
			    {Methods}&NE$\downarrow$&SR$\uparrow$&nDTW$\uparrow$&sDTW$\uparrow$ &NE$\downarrow$&SR$\uparrow$&nDTW$\uparrow$&sDTW$\uparrow$ \\
				\midrule
				{LSTM~\cite{ku2020room}}  &10.7  &25.2 &42.2 &20.7   &10.9&22.8&38.9&18.2\\
                {EnvDrop+~\cite{shenmuch}}    &-  &-  &-  &- &- &43.6&55.7&-\\
                {HAMT~\cite{chen2021history}}    &-  &59.4  &65.3  &50.9 &- &56.5&63.1&48.3\\
                {EnvEdit~\cite{li2022envedit}}   &-  &67.2  &71.1  &58.5 &- &62.8&68.5&54.6\\
			{MPM~\cite{dou2023masked}}    &-  &67.7  &71.0  &58.9 &- &63.5&67.7&54.5\\
                {MARVEL~\cite{wang2022less}}   &3.0  &75.9  &79.1  &68.8 &4.5 &64.8&70.8&57.5\\
                {BEVBert~\cite{an2023bevbert}}    &3.2  &75.0  &76.3  &66.7 &4.0 &68.5&69.6&58.6\\
                {PRET~\cite{lu2024pret}}    &2.4  &79.3  &80.4  &70.7 &3.2 &72.8&73.4&62.4\\
				\midrule
                {FIGR}    &\textbf{1.8}&\textbf{84.1}&\textbf{84.7}&\textbf{76.2}&\textbf{2.7}&\textbf{75.1}&\textbf{75.3}&\textbf{64.8}\\
				\bottomrule
			\end{tabular}}
	        \label{tab:rxr}
	\end{center}
\end{table}
\subsubsection{Evaluation Metrics}
Due to the non-shortest-path nature of trajectories in RxR, trajectory length–based metrics such as TL and SPL are not applicable. Following established evaluation protocols~\cite{an2023bevbert,lu2024pret}, we report four metrics:  Navigation Error ($\textit{NE}$), Success Rate ($\textit{SR}$),  Normalized Dynamic Time Warping ($\textit{nDTW}$), and  Success weighted Dynamic Time Warping ($\textit{sDTW}$). Among these, $\textit{nDTW}$ measures the similarity between the predicted and ground-truth trajectories, while $\textit{sDTW}$ further incorporates task success, providing a comprehensive assessment of trajectory alignment quality.

\subsubsection{Quantitative Results}
\begin{figure}[t]
	\centerline{\includegraphics[width=1.0\linewidth]{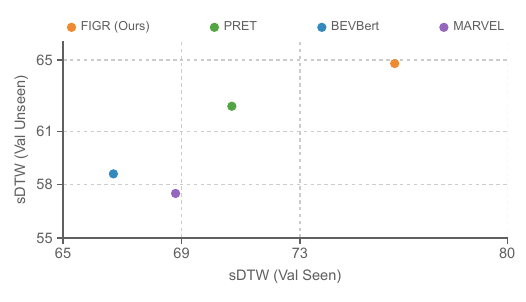}}
    \caption{  
Comparison of validation-seen and validation-unseen sDTW for FIGR, PRET, BEVBert, and MARVEL on RxR using the results in Table~\ref{tab:rxr}. Higher values are better on both axes.
    }
	\label{fig:figure1}
\end{figure}

We compare the proposed method with several representative baselines, including LSTM-based models~\cite{ku2020room}, EnvDrop+~\cite{shenmuch}, HAMT~\cite{chen2021history}, EnvEdit~\cite{li2022envedit}, MPM~\cite{dou2023masked}, MARVEL~\cite{wang2022less}, BEVBert~\cite{an2023bevbert}, and PRET~\cite{lu2024pret}. Quantitative results on the RxR benchmark are reported in Table~\ref{tab:rxr}.
On validation seen, FIGR achieved an SR of 84.1\% and an sDTW of 76.2, exceeding PRET by 4.8 percentage points and 5.5 points, respectively. On validation unseen, FIGR attained an SR of 75.1\%, an nDTW of 75.3, and an sDTW of 64.8. Relative to PRET, these correspond to gains of 2.3 percentage points in SR and 1.9 and 2.4 points in nDTW and sDTW, respectively. NE also decreased from 3.2\,m to 2.7\,m on this split. These results indicate improvements in both endpoint accuracy and trajectory alignment in unseen environments; Fig.~\ref{fig:figure1} visualizes the sDTW comparison across splits.
\begin{table}[h]
\centering
\caption{Ablation of OGI on R2R validation unseen with full AOVG. D: angular--visual decoupling; GE: geometric embedding. O5 replaces the interaction decoder with an MLP. NE is in meters; SR and SPL use a 0--100 scale.}
\label{tab:ablation}
\small
\setlength{\tabcolsep}{4pt}
\begin{tabular}{lccrrr}
\toprule
ID & D & GE & NE$\downarrow$ & SR$\uparrow$ & SPL$\uparrow$\\
\midrule
O1 & -- & -- & 2.88 & 73.4 & 64.1\\
O2 & $\checkmark$ & -- & 2.78 & 74.5 & 65.0\\
O3 & -- & $\checkmark$ & 2.73 & 74.8 & 65.5\\
O4 (full) & $\checkmark$ & $\checkmark$ & 2.61 & 76.0 & 67.0\\
\midrule
O5 (MLP) & $\checkmark$ & $\checkmark$ & 2.83 & 74.0 & 64.9\\
\bottomrule
\end{tabular}
\end{table}

\subsection{Ablation Study}
\label{sec:ablation}

We conduct ablation studies to examine the contributions of observation--graph interaction, adaptive open-vocabulary guidance, and multilingual adaptation. Tables~\ref{tab:ablation}--\ref{tab:msa_ablation} separate component-level effects from vocabulary coverage, compositional reasoning, and language-specific performance.

\subsubsection{Evaluation Protocol}
All comparisons are reported on the validation-unseen splits. Unless otherwise specified, components outside the target ablation remain active. R2R variants bypass MSA, whereas RxR variants retain full MSA except in the dedicated adapter comparison. We report NE, SR, and SPL for the Observation-Graph Interaction (OGI) analysis, navigation and subset-level metrics for AOVG, and language-specific SR, nDTW, and sDTW for MSA.

\subsubsection{Observation--Graph Interaction}
Table~\ref{tab:ablation} uses a $2\times2$ design for angular--visual decoupling (D) and geometric embedding (GE), with AOVG fixed. Disabling D replaces the separate angular and visual embedding branches with a joint MLP of matched output width. Disabling GE sets $pe=0$ while preserving the heading/elevation encoding $\mathcal C(\cdot)$. The candidate-query/observation-key-value decoder remains active in O1--O4. O5 replaces this decoder with an MLP applied to each candidate feature concatenated with the pooled observation, testing direction-selective interaction without removing visual input.
\begin{table*}[!t]
\centering
\caption{AOVG ablations on R2R and RxR validation unseen with OGI retained and MSA active for RxR. The last two columns report sDTW on the RxR OOV and compositional-relation subsets. All metrics use a 0--100 scale.}
\label{tab:extension_ablation}
\small
\setlength{\tabcolsep}{7pt}
\begin{tabular}{llrrrrrr}
\toprule
 & & \multicolumn{2}{c}{R2R} & \multicolumn{2}{c}{RxR (aggregate)} & \multicolumn{2}{c}{RxR subsets}\\
\cmidrule(lr){3-4}\cmidrule(lr){5-6}\cmidrule(lr){7-8}
ID & Variant & SR$\uparrow$ & SPL$\uparrow$ & SR$\uparrow$ & sDTW$\uparrow$ & OOV$\uparrow$ & Compositional$\uparrow$\\
\midrule
A0 & No key-detail guidance & 72.8 & 63.5 & 70.9 & 59.7 & 57.7 & 55.8\\
A1 & Fixed-vocabulary KDG & 74.8 & 65.3 & 72.6 & 61.7 & 57.2 & 57.4\\
\midrule
A2 & No parser state conditioning & 74.6 & 65.4 & 73.2 & 62.3 & 61.4 & 59.2\\
A3 & No location-role channel & 74.3 & 65.0 & 72.9 & 61.9 & 60.7 & 59.6\\
A4 & No object-role channel & 74.5 & 65.2 & 73.3 & 62.4 & 59.8 & 60.3\\
A5 & No spatial-relation channel & 74.1 & 64.9 & 72.7 & 61.7 & 61.2 & 57.8\\
A6 & No other-role channel & 75.8 & 66.8 & 74.9 & 64.6 & 64.1 & 62.8\\
A7 & No semantic-graph encoding & 74.7 & 65.5 & 73.5 & 62.6 & 62.3 & 59.0\\
A8 & No confidence gate & 76.1 & 66.7 & 75.3 & 64.5 & 60.6 & 62.5\\
A9 & No role/relation/grounding losses & 74.8 & 65.5 & 73.6 & 62.8 & 61.0 & 60.2\\
\midrule
A10 & Full AOVG & 76.0 & 67.0 & 75.1 & 64.8 & 64.5 & 63.0\\
\bottomrule
\end{tabular}
\end{table*}

Angular--visual decoupling and geometric embedding improve navigation performance both independently and jointly. Relative to O1, D alone and GE alone increase SPL by 0.9 and 1.4 points, respectively. Combining them raises SPL from 64.1 to 67.0 and SR from 73.4 to 76.0, while reducing NE from 2.88 to 2.61~m. The descriptive interaction contrast, 0.6 SPL points, is consistent with complementary effects of the two components. Replacing the interaction decoder with an MLP reduces SPL by 2.1 points relative to O4, supporting candidate-conditioned observation interaction beyond feature encoding alone.

\subsection{Dynamic and Multilingual Analysis}
\label{sec:dynamic_multilingual_analysis}

\subsubsection{AOVG Components and Diagnostic Subsets}
Table~\ref{tab:extension_ablation} compares AOVG variants with OGI retained. A0 disables key-detail enhancement by setting $F_e^{(t)}=F_c^{(t)}$ and removing AOVG-specific losses. A1 replaces contextual parsing with fixed location/object dictionaries and exact matching, without semantic-graph encoding. The RxR dictionaries are language-specific and restricted to training instructions. A0 and A1 are reference systems rather than parameter-matched single-factor ablations.

A2 removes state conditioning by supplying a zero query to the span-attention, role, and relation predictors, while leaving the confidence gate state-conditioned. A3--A6 individually zero the corresponding role embedding before fusion, retaining the parser, graph, and losses to isolate each direct role channel. A7 replaces only $\operatorname{GNN}(\mathcal S_t)$ with a zero vector, retaining all four role embeddings. Thus, A5 removes the direct spatial-relation channel, whereas A7 removes explicit graph encoding; neither removes all relational information. A8 fixes $g_t=\mathbf{1}$. A9 sets $\lambda_r=\lambda_s=\lambda_g=0$, retaining the complete architecture and, on RxR, $\mathcal L_{\mathrm{msa}}$. This last comparison tests the auxiliary objectives jointly, not their individual effects.

The RxR OOV subset contains instructions with at least one location/object expression absent from the fixed training dictionary of the corresponding language. The compositional-relation subset contains instructions linking an entity to a directional or ordinal constraint. These subsets are not mutually exclusive. Their sDTW scores distinguish performance on unfamiliar expressions and compositional constraints from aggregate navigation performance.

Full AOVG improves R2R SPL from 63.5 to 67.0 and RxR sDTW from 59.7 to 64.8 relative to the no-guidance baseline. Compared with fixed-vocabulary KDG, the gains are 1.7 SPL points on R2R and 3.1 sDTW points on RxR. The OOV sDTW gain is larger, reaching 7.3 points, which supports the use of contextual span representations for expressions outside the fixed vocabulary.

The component comparisons further distinguish contextual selection from relational structure. Removing parser state conditioning reduces RxR sDTW by 2.5 points. Removing the location and object channels reduces this metric by 2.9 and 2.4 points, respectively. The spatial-relation channel and semantic graph are particularly relevant to compositional instructions: their removal lowers compositional-subset sDTW by 5.2 and 4.0 points, respectively. These results support retaining both role-specific features and explicit graph encoding.

Confidence gating improves path fidelity without maximizing aggregate SR in every comparison. The gate-free variant achieves slightly higher SR on R2R and RxR, but lower SPL on R2R and lower sDTW on RxR; its OOV sDTW decreases from 64.5 to 60.6. Removing the other-role channel produces smaller differences of 0.2 points in R2R SPL and RxR sDTW. Removing the role, relation, and grounding objectives jointly reduces RxR sDTW by 2.0 points, supporting their combined contribution without isolating the effect of each loss.

\subsubsection{Multilingual Semantic Adapter}
Table~\ref{tab:msa_ablation} compares adapter designs with OGI, the multilingual language encoder, and full AOVG retained. M0 bypasses MSA and omits $\mathcal L_{\mathrm{msa}}$. M1 uses a shared residual adapter of width $d_a$ without a language-specific bias. M2 increases the shared adapter width to approximately match the trainable parameter count of the language-specific adapters and bias projections. M3 retains language-specific MSA but sets $\lambda_m=0$, whereas M4 uses the complete MSA. The role and relation heads remain shared across languages. These comparisons distinguish language conditioning from adapter capacity and alignment supervision.

\begin{table*}[t]
\centering
\caption{MSA ablations on RxR validation unseen with OGI and full AOVG retained. Results are reported separately for English, Hindi, and Telugu. Macro sDTW denotes the equal-weight language mean rather than the episode-weighted aggregate. All metrics use a 0--100 scale.}
\label{tab:msa_ablation}
\small
\setlength{\tabcolsep}{4pt}
\begin{tabular}{llrrrrrrrrrr}
\toprule
 & & \multicolumn{3}{c}{English} & \multicolumn{3}{c}{Hindi} & \multicolumn{3}{c}{Telugu} & Macro\\
\cmidrule(lr){3-5}\cmidrule(lr){6-8}\cmidrule(lr){9-11}
ID & Variant & SR$\uparrow$ & nDTW$\uparrow$ & sDTW$\uparrow$ & SR$\uparrow$ & nDTW$\uparrow$ & sDTW$\uparrow$ & SR$\uparrow$ & nDTW$\uparrow$ & sDTW$\uparrow$ & sDTW$\uparrow$\\
\midrule
M0 & No MSA & 76.6 & 76.5 & 66.5 & 72.4 & 72.6 & 61.7 & 69.9 & 71.2 & 59.4 & 62.5\\
M1 & Shared adapter & 77.0 & 76.9 & 66.9 & 73.4 & 73.7 & 62.7 & 71.0 & 72.0 & 60.4 & 63.3\\
M2 & Shared, capacity-matched & 77.2 & 77.0 & 67.0 & 73.7 & 74.0 & 63.0 & 71.5 & 72.5 & 61.1 & 63.7\\
M3 & MSA, $\lambda_m=0$ & 77.3 & 77.1 & 67.1 & 74.1 & 74.4 & 63.7 & 72.2 & 73.0 & 61.8 & 64.2\\
\midrule
M4 & Full MSA & 77.4 & 77.2 & 67.3 & 74.8 & 75.0 & 64.5 & 73.1 & 73.7 & 62.6 & 64.8\\
\bottomrule
\end{tabular}
\end{table*}

Full MSA improves sDTW over the no-adapter baseline by 0.8, 2.8, and 3.2 points for English, Hindi, and Telugu, respectively, while also increasing SR and nDTW in all three languages. The larger gains for Hindi and Telugu indicate that the benefit of adaptation varies across the evaluated languages. Compared with the capacity-matched shared adapter, full MSA improves sDTW by 0.3, 1.5, and 1.5 points, respectively, suggesting that additional parameter capacity alone does not account for the observed gains.

The alignment objective provides a further contribution: relative to M3, full MSA raises English, Hindi, and Telugu sDTW by 0.2, 0.8, and 0.8 points, respectively. Together, the adapter comparisons support language-conditioned residual transformations and alignment supervision for multilingual navigation. 

\section{CONCLUSION}
We presented FIGR, a vision-and-language navigation framework combining observation--graph interaction, adaptive open-vocabulary guidance, and multilingual semantic adaptation. Experiments on R2R and RxR show improved path efficiency and trajectory alignment relative to the compared baselines, while ablation studies support the contributions of the proposed components. The theoretical analysis provides complementary insights under its stated assumptions. Remaining challenges include uncertain grounding of rare or implicit expressions, generalization beyond the three RxR languages, and transfer from simulation to real-world environments. Future work will address these limitations through confidence-aware grounding, broader multilingual evaluation, and continuous real-world navigation.

  \ifCLASSOPTIONcaptionsoff
  \newpage
  \fi

  \bibliographystyle{ieeetr}
  \bibliography{ref}

@inproceedings{chang2017matterport3d,
  title={Matterport3D: Learning from RGB-D Data in Indoor Environments},
  author={Chang, Angel and Dai, Angela and Funkhouser, Thomas and Halber, Maciej and Niebner, Matthias and Savva, Manolis and Song, Shuran and Zeng, Andy and Zhang, Yinda},
  booktitle={International Conference on 3D Vision (3DV)},
  year={2017}
}

@inproceedings{anderson2018vision,
  title={Vision-and-language navigation: Interpreting visually-grounded navigation instructions in real environments},
  author={Anderson, Peter and Wu, Qi and Teney, Damien and Bruce, Jake and Johnson, Mark and S{\"u}nderhauf, Niko and Reid, Ian and Gould, Stephen and Van Den Hengel, Anton},
  booktitle={Proceedings of the IEEE Conference on Computer Vision and Pattern Recognition},
  pages={3674--3683},
  year={2018}
}

@inproceedings{ku2020room,
  title={Room-Across-Room: Multilingual Vision-and-Language Navigation with Dense Spatiotemporal Grounding},
  author={Ku, Alexander and Anderson, Peter and Patel, Roma and Ie, Eugene and Baldridge, Jason},
  booktitle={Proceedings of the 2020 Conference on Empirical Methods in Natural Language Processing (EMNLP)},
  pages={4392--4412},
  year={2020}
}

@inproceedings{xie2025vln,
  title={Observation-Graph Interaction and Key-Detail Guidance for Vision and Language Navigation},
  author={Yifan Xie; Binkai Ou; Fei Ma; Yaohua Liu},
  booktitle={IEEE/RSJ International Conference on Intelligent Robots and Systems (IROS)},
  pages={15306--15313},
  year={2025}
}

@inproceedings{chen2022think,
  title={Think global, act local: Dual-scale graph transformer for vision-and-language navigation},
  author={Chen, Shizhe and Guhur, Pierre-Louis and Tapaswi, Makarand and Schmid, Cordelia and Laptev, Ivan},
  booktitle={Proceedings of the IEEE/CVF Conference on Computer Vision and Pattern Recognition},
  pages={16537--16547},
  year={2022}
}

@article{oquab2024dinov2,
  title={DINOv2: Learning Robust Visual Features without Supervision},
  author={Oquab, Maxime and Darcet, Timoth{\'e}e and Moutakanni, Th{\'e}o and Vo, Huy and Szafraniec, Marc and Khalidov, Vasil and Fernandez, Pierre and Haziza, Daniel and Massa, Francisco and El-Nouby, Alaaeldin and others},
  journal={Transactions on Machine Learning Research Journal},
  pages={1--31},
  year={2024}
}

@inproceedings{an2023bevbert,
  title={Bevbert: Multimodal map pre-training for language-guided navigation},
  author={An, Dong and Qi, Yuankai and Li, Yangguang and Huang, Yan and Wang, Liang and Tan, Tieniu and Shao, Jing},
  booktitle={Proceedings of the IEEE/CVF International Conference on Computer Vision},
  pages={2737--2748},
  year={2023}
}

@article{xie2024hecpg,
  title={HECPG: Hyperbolic Embedding and Confident Patch-Guided Network for Point Cloud Matching},
  author={Xie, Yifan and Zhu, Jihua and Li, Shiqi and Hu, Naiwen and Shi, Pengcheng},
  journal={IEEE Transactions on Geoscience and Remote Sensing},
  year={2024},
  publisher={IEEE}
}

@inproceedings{chen2021topological,
  title={Topological planning with transformers for vision-and-language navigation},
  author={Chen, Kevin and Chen, Junshen K and Chuang, Jo and V{\'a}zquez, Marynel and Savarese, Silvio},
  booktitle={Proceedings of the IEEE/CVF Conference on Computer Vision and Pattern Recognition},
  pages={11276--11286},
  year={2021}
}

@article{vaswani2017attention,
  title={Attention is all you need},
  author={Vaswani, A},
  journal={Advances in Neural Information Processing Systems},
  year={2017}
}

@inproceedings{lu2024pret,
  title={Pret: Planning with directed fidelity trajectory for vision and language navigation},
  author={Lu, Renjie and Meng, Jingke and Zheng, Wei-Shi},
  booktitle={European Conference on Computer Vision},
  pages={72--88},
  year={2024},
  organization={Springer}
}

@article{xie2024pointtalk,
  title={PointTalk: Audio-Driven Dynamic Lip Point Cloud for 3D Gaussian-based Talking Head Synthesis},
  author={Xie, Yifan and Feng, Tao and Zhang, Xin and Luo, Xiangyang and Guo, Zixuan and Yu, Weijiang and Chang, Heng and Ma, Fei and Yu, Fei Richard},
  journal={arXiv preprint arXiv:2412.08504},
  year={2024}
}

@article{li2021align,
  title={Align before fuse: Vision and language representation learning with momentum distillation},
  author={Li, Junnan and Selvaraju, Ramprasaath and Gotmare, Akhilesh and Joty, Shafiq and Xiong, Caiming and Hoi, Steven Chu Hong},
  journal={Advances in Neural Information Processing Systems},
  volume={34},
  pages={9694--9705},
  year={2021}
}

@article{conneau2019unsupervised,
  title={Unsupervised cross-lingual representation learning at scale},
  author={Conneau, A},
  journal={arXiv preprint arXiv:1911.02116},
  year={2019}
}

@inproceedings{hao2020towards,
  title={Towards learning a generic agent for vision-and-language navigation via pre-training},
  author={Hao, Weituo and Li, Chunyuan and Li, Xiujun and Carin, Lawrence and Gao, Jianfeng},
  booktitle={Proceedings of the IEEE/CVF Conference on Computer Vision and Pattern Recognition},
  pages={13137--13146},
  year={2020}
}

@article{paszke2019pytorch,
  title={Pytorch: An imperative style, high-performance deep learning library},
  author={Paszke, Adam and Gross, Sam and Massa, Francisco and Lerer, Adam and Bradbury, James and Chanan, Gregory and Killeen, Trevor and Lin, Zeming and Gimelshein, Natalia and Antiga, Luca and others},
  journal={Advances in Neural Information Processing Systems},
  volume={32},
  year={2019}
}

@article{loshchilov2017decoupled,
  title={Decoupled weight decay regularization},
  author={Loshchilov, I},
  journal={arXiv preprint arXiv:1711.05101},
  year={2017}
}

@inproceedings{wang2022less,
  title={Less is more: Generating grounded navigation instructions from landmarks},
  author={Wang, Su and Montgomery, Ceslee and Orbay, Jordi and Birodkar, Vighnesh and Faust, Aleksandra and Gur, Izzeddin and Jaques, Natasha and Waters, Austin and Baldridge, Jason and Anderson, Peter},
  booktitle={Proceedings of the IEEE/CVF Conference on Computer Vision and Pattern Recognition},
  pages={15428--15438},
  year={2022}
}

@article{fried2018speaker,
  title={Speaker-follower models for vision-and-language navigation},
  author={Fried, Daniel and Hu, Ronghang and Cirik, Volkan and Rohrbach, Anna and Andreas, Jacob and Morency, Louis-Philippe and Berg-Kirkpatrick, Taylor and Saenko, Kate and Klein, Dan and Darrell, Trevor},
  journal={Advances in neural information processing systems},
  volume={31},
  year={2018}
}

@inproceedings{wang2019reinforced,
  title={Reinforced cross-modal matching and self-supervised imitation learning for vision-language navigation},
  author={Wang, Xin and Huang, Qiuyuan and Celikyilmaz, Asli and Gao, Jianfeng and Shen, Dinghan and Wang, Yuan-Fang and Wang, William Yang and Zhang, Lei},
  booktitle={Proceedings of the IEEE/CVF conference on computer vision and pattern recognition},
  pages={6629--6638},
  year={2019}
}

@inproceedings{ma2019regretful,
  title={The regretful agent: Heuristic-aided navigation through progress estimation},
  author={Ma, Chih-Yao and Wu, Zuxuan and AlRegib, Ghassan and Xiong, Caiming and Kira, Zsolt},
  booktitle={Proceedings of the IEEE/CVF conference on Computer Vision and Pattern Recognition},
  pages={6732--6740},
  year={2019}
}

@article{xie2023cross,
  title={Cross-modal information-guided network using contrastive learning for point cloud registration},
  author={Xie, Yifan and Zhu, Jihua and Li, Shiqi and Shi, Pengcheng},
  journal={IEEE Robotics and Automation Letters},
  volume={9},
  number={1},
  pages={103--110},
  year={2023},
  publisher={IEEE}
}

@inproceedings{tan2019learning,
  title={Learning to Navigate Unseen Environments: Back Translation with Environmental Dropout},
  author={Tan, Hao and Yu, Licheng and Bansal, Mohit},
  booktitle={Proceedings of the 2019 Conference of the North American Chapter of the Association for Computational Linguistics: Human Language Technologies, Volume 1 (Long and Short Papers)},
  pages={2610--2621},
  year={2019}
}

@inproceedings{an2021neighbor,
  title={Neighbor-view enhanced model for vision and language navigation},
  author={An, Dong and Qi, Yuankai and Huang, Yan and Wu, Qi and Wang, Liang and Tan, Tieniu},
  booktitle={Proceedings of the 29th ACM International Conference on Multimedia},
  pages={5101--5109},
  year={2021}
}

@inproceedings{wang2021structured,
  title={Structured scene memory for vision-language navigation},
  author={Wang, Hanqing and Wang, Wenguan and Liang, Wei and Xiong, Caiming and Shen, Jianbing},
  booktitle={Proceedings of the IEEE/CVF conference on Computer Vision and Pattern Recognition},
  pages={8455--8464},
  year={2021}
}

@inproceedings{hong2021vln,
  title={Vln bert: A recurrent vision-and-language bert for navigation},
  author={Hong, Yicong and Wu, Qi and Qi, Yuankai and Rodriguez-Opazo, Cristian and Gould, Stephen},
  booktitle={Proceedings of the IEEE/CVF conference on Computer Vision and Pattern Recognition},
  pages={1643--1653},
  year={2021}
}

@article{chen2021history,
  title={History aware multimodal transformer for vision-and-language navigation},
  author={Chen, Shizhe and Guhur, Pierre-Louis and Schmid, Cordelia and Laptev, Ivan},
  journal={Advances in neural information processing systems},
  volume={34},
  pages={5834--5847},
  year={2021}
}

@inproceedings{lin2022multimodal,
  title={Multimodal transformer with variable-length memory for vision-and-language navigation},
  author={Lin, Chuang and Jiang, Yi and Cai, Jianfei and Qu, Lizhen and Haffari, Gholamreza and Yuan, Zehuan},
  booktitle={European Conference on Computer Vision},
  pages={380--397},
  year={2022},
  organization={Springer}
}

@inproceedings{chaplot2020neural,
  title={Neural topological slam for visual navigation},
  author={Chaplot, Devendra Singh and Salakhutdinov, Ruslan and Gupta, Abhinav and Gupta, Saurabh},
  booktitle={Proceedings of the IEEE/CVF conference on computer vision and pattern recognition},
  pages={12875--12884},
  year={2020}
}

@article{chaplot2020object,
  title={Object goal navigation using goal-oriented semantic exploration},
  author={Chaplot, Devendra Singh and Gandhi, Dhiraj Prakashchand and Gupta, Abhinav and Salakhutdinov, Russ R},
  journal={Advances in Neural Information Processing Systems},
  volume={33},
  pages={4247--4258},
  year={2020}
}

@inproceedings{henriques2018mapnet,
  title={Mapnet: An allocentric spatial memory for mapping environments},
  author={Henriques, Joao F and Vedaldi, Andrea},
  booktitle={proceedings of the IEEE Conference on Computer Vision and Pattern Recognition},
  pages={8476--8484},
  year={2018}
}

@article{wang2024survey,
  title={A survey of visual SLAM in dynamic environment: the evolution from geometric to semantic approaches},
  author={Wang, Yanan and Tian, Yaobin and Chen, Jiawei and Xu, Kun and Ding, Xilun},
  journal={IEEE Transactions on Instrumentation and Measurement},
  year={2024},
  publisher={IEEE}
}

@article{fuentes2015visual,
  title={Visual simultaneous localization and mapping: a survey},
  author={Fuentes-Pacheco, Jorge and Ruiz-Ascencio, Jos{\'e} and Rend{\'o}n-Mancha, Juan Manuel},
  journal={Artificial intelligence review},
  volume={43},
  pages={55--81},
  year={2015},
  publisher={Springer}
}

@inproceedings{gao2023adaptive,
  title={Adaptive zone-aware hierarchical planner for vision-language navigation},
  author={Gao, Chen and Peng, Xingyu and Yan, Mi and Wang, He and Yang, Lirong and Ren, Haibing and Li, Hongsheng and Liu, Si},
  booktitle={Proceedings of the IEEE/CVF Conference on Computer Vision and Pattern Recognition},
  pages={14911--14920},
  year={2023}
}

@inproceedings{hwang2023meta,
  title={Meta-explore: Exploratory hierarchical vision-and-language navigation using scene object spectrum grounding},
  author={Hwang, Minyoung and Jeong, Jaeyeon and Kim, Minsoo and Oh, Yoonseon and Oh, Songhwai},
  booktitle={Proceedings of the IEEE/CVF Conference on Computer Vision and Pattern Recognition},
  pages={6683--6693},
  year={2023}
}

@inproceedings{wang2023gridmm,
  title={Gridmm: Grid memory map for vision-and-language navigation},
  author={Wang, Zihan and Li, Xiangyang and Yang, Jiahao and Liu, Yeqi and Jiang, Shuqiang},
  booktitle={Proceedings of the IEEE/CVF International Conference on Computer Vision},
  pages={15625--15636},
  year={2023}
}

@inproceedings{shenmuch,
  title={How Much Can CLIP Benefit Vision-and-Language Tasks?},
  author={Shen, Sheng and Li, Liunian Harold and Tan, Hao and Bansal, Mohit and Rohrbach, Anna and Chang, Kai-Wei and Yao, Zhewei and Keutzer, Kurt},
  booktitle={International Conference on Learning Representations},
  year={2022},
  publisher={OpenReview.net}
}

@inproceedings{li2022envedit,
  title={Envedit: Environment editing for vision-and-language navigation},
  author={Li, Jialu and Tan, Hao and Bansal, Mohit},
  booktitle={Proceedings of the IEEE/CVF Conference on Computer Vision and Pattern Recognition},
  pages={15407--15417},
  year={2022}
}

@inproceedings{dou2023masked,
  title={Masked Path Modeling for Vision-and-Language Navigation},
  author={Dou, Zi-Yi and Gao, Feng and Peng, Nanyun},
  booktitle={Findings of the Association for Computational Linguistics: EMNLP 2023},
  pages={15255--15269},
  year={2023},
  address={Singapore},
  publisher={Association for Computational Linguistics}
}

@inproceedings{gu2022vision,
  title={Vision-and-Language Navigation: A Survey of Tasks, Methods, and Future Directions},
  author={Gu, Jing and Stefani, Eliana and Wu, Qi and Thomason, Jesse and Wang, Xin},
  booktitle={Proceedings of the 60th Annual Meeting of the Association for Computational Linguistics (Volume 1: Long Papers)},
  pages={7606--7623},
  year={2022}
}

@inproceedings{radford2021learning,
  title={Learning transferable visual models from natural language supervision},
  author={Radford, Alec and Kim, Jong Wook and Hallacy, Chris and Ramesh, Aditya and Goh, Gabriel and Agarwal, Sandhini and Sastry, Girish and Askell, Amanda and Mishkin, Pamela and Clark, Jack and others},
  booktitle={International conference on machine learning},
  pages={8748--8763},
  year={2021},
  organization={PMLR}
}

@inproceedings{liu2021vision,
  title={Vision-language navigation with random environmental mixup},
  author={Liu, Chong and Zhu, Fengda and Chang, Xiaojun and Liang, Xiaodan and Ge, Zongyuan and Shen, Yi-Dong},
  booktitle={Proceedings of the IEEE/CVF International Conference on Computer Vision},
  pages={1644--1654},
  year={2021}
}

@article{lin2025navcot,
  title={Navcot: Boosting llm-based vision-and-language navigation via learning disentangled reasoning},
  author={Lin, Bingqian and Nie, Yunshuang and Wei, Ziming and Chen, Jiaqi and Ma, Shikui and Han, Jianhua and Xu, Hang and Chang, Xiaojun and Liang, Xiaodan},
  journal={IEEE Transactions on Pattern Analysis and Machine Intelligence},
  year={2025},
  publisher={IEEE}
}

@article{yu2025mossvln,
  title={MossVLN: Memory-Observation Synergistic System for Continuous Vision-Language Navigation},
  author={Yu, Ting and Wu, Yifei and Cui, Qiongjie and Huang, Qingming and Yu, Jun},
  journal={IEEE Transactions on Multimedia},
  year={2025},
  publisher={IEEE}
}

@article{chen2025constraint,
  title={Constraint-aware zero-shot vision-language navigation in continuous environments},
  author={Chen, Kehan and An, Dong and Huang, Yan and Xu, Rongtao and Su, Yifei and Ling, Yonggen and Reid, Ian and Wang, Liang},
  journal={IEEE Transactions on Pattern Analysis and Machine Intelligence},
  year={2025},
  publisher={IEEE}
}

@article{wen2025ovl,
  title={OVL-MAP: An Online Visual Language Map Approach for Vision-and-Language Navigation in Continuous Environments},
  author={Wen, Shuhuan and Zhang, Ziyuan and Sun, Yuxiang and Wang, Zhiwen},
  journal={IEEE Robotics and Automation Letters},
  year={2025},
  publisher={IEEE}
}

@inproceedings{liu2024volumetric,
  title={Volumetric environment representation for vision-language navigation},
  author={Liu, Rui and Wang, Wenguan and Yang, Yi},
  booktitle={Proceedings of the IEEE/CVF conference on computer vision and pattern recognition},
  pages={16317--16328},
  year={2024}
}

@article{wu2024vision,
  title={Vision-and-language navigation via latent semantic alignment learning},
  author={Wu, Siying and Fu, Xueyang and Wu, Feng and Zha, Zheng-Jun},
  journal={IEEE Transactions on Multimedia},
  volume={26},
  pages={8406--8418},
  year={2024},
  publisher={IEEE}
}

@article{wang2023skill,
  title={Skill-based hierarchical reinforcement learning for target visual navigation},
  author={Wang, Shuo and Wu, Zhihao and Hu, Xiaobo and Lin, Youfang and Lv, Kai},
  journal={IEEE Transactions on Multimedia},
  volume={25},
  pages={8920--8932},
  year={2023},
  publisher={IEEE}
}

@article{qiao2023hop+,
  title={Hop+: History-enhanced and order-aware pre-training for vision-and-language navigation},
  author={Qiao, Yanyuan and Qi, Yuankai and Hong, Yicong and Yu, Zheng and Wang, Peng and Wu, Qi},
  journal={IEEE Transactions on Pattern Analysis and Machine Intelligence},
  volume={45},
  number={7},
  pages={8524--8537},
  year={2023},
  publisher={IEEE}
}

@inproceedings{liu2023bird,
  title={Bird's-eye-view scene graph for vision-language navigation},
  author={Liu, Rui and Wang, Xiaohan and Wang, Wenguan and Yang, Yi},
  booktitle={Proceedings of the IEEE/CVF International Conference on Computer Vision},
  pages={10968--10980},
  year={2023}
}

@inproceedings{li2023kerm,
  title={Kerm: Knowledge enhanced reasoning for vision-and-language navigation},
  author={Li, Xiangyang and Wang, Zihan and Yang, Jiahao and Wang, Yaowei and Jiang, Shuqiang},
  booktitle={Proceedings of the IEEE/CVF Conference on Computer Vision and Pattern Recognition},
  pages={2583--2592},
  year={2023}
}

@inproceedings{wang2023lana,
  title={Lana: A language-capable navigator for instruction following and generation},
  author={Wang, Xiaohan and Wang, Wenguan and Shao, Jiayi and Yang, Yi},
  booktitle={Proceedings of the IEEE/CVF conference on computer vision and pattern recognition},
  pages={19048--19058},
  year={2023}
}

@inproceedings{li2023improving,
  title={Improving vision-and-language navigation by generating future-view image semantics},
  author={Li, Jialu and Bansal, Mohit},
  booktitle={Proceedings of the IEEE/CVF conference on computer vision and pattern recognition},
  pages={10803--10812},
  year={2023}
}

@inproceedings{lin2023learning,
  title={Learning vision-and-language navigation from youtube videos},
  author={Lin, Kunyang and Chen, Peihao and Huang, Diwei and Li, Thomas H and Tan, Mingkui and Gan, Chuang},
  booktitle={Proceedings of the IEEE/CVF International Conference on Computer Vision},
  pages={8317--8326},
  year={2023}
}

@article{yuan2026spenav,
  title={SPENav: Dynamic Object Filtering With Spatial Perception Enhancement for Vision-Language Navigation},
  author={Yuan, Shuai and Zhang, Huaxiang and Liu, Li and Zhu, Lei and Dong, Xinfeng},
  journal={IEEE Transactions on Circuits and Systems for Video Technology},
  year={2026},
  publisher={IEEE}
}

\section*{Supplementary material I: Proof of Theorem~\ref{thm:ogi}}\label{sec:proof:1}

\begin{proof}
Let $g_o$ and $g_g$ denote the gradients propagated through the observation and graph branches, respectively. We compare the coupled representation with the representation produced by the OGI operations in Sec.~\ref{sec:ogi}; the result concerns the navigation objective $\mathcal L_{\mathrm{nav}}$ only. Assume that the observation maps are differentiable, that the residual cross-channel interference has zero mean and covariance matrices $\Sigma_a,\Sigma_v\succeq0$, and that the decoder and geometric embedding are Lipschitz.

Define the loss gradient with respect to the original coupled features \(F_o\) as
\begin{equation}
\nabla_{F_o} \mathcal{L}_{\mathrm{nav}} = \frac{\partial \mathcal{L}_{\mathrm{nav}}}{\partial F_o}.
\end{equation}
By the chain rule,
\begin{equation}\label{eq:20}
\nabla_{F_o} \mathcal{L}_{\mathrm{nav}} = g_a P_a + g_v P_v + \xi_o,
\end{equation}
where $P_a$ and $P_v$ are the selection Jacobians of the angular and visual channels and $\xi_o$ is the residual interference term. Unlike an identity-Jacobian assumption, this form is valid for the channel decomposition used in Method.
The residual term contributes
\begin{equation}
\mathbb E\|\nabla_{F_o}\mathcal L_{\mathrm{nav}}\|^2
=\mathbb E\|g_aP_a+g_vP_v\|^2+\mathbb E\|\xi_o\|^2,
\end{equation}
under the zero-mean orthogonality assumption. For the decoupled branch, the corresponding term is absent, so the observation-side reduction is $\delta_o=\mathbb E\|\xi_o\|^2\ge0$.

For the graph branch, define $J_g$ and $J'_g$ as the Jacobians of the candidate features before and after adding the heading--elevation positional embedding and the candidate-query/observation-key-value decoder. Lipschitz continuity gives
\begin{equation}
\left|\mathbb E\|g_gJ'_g\|^2-\mathbb E\|g_gJ_g\|^2\right|\le \gamma_g
\end{equation}
for a finite $\gamma_g\ge0$. The cross term between the observation and graph gradients is bounded by Cauchy--Schwarz, and we denote its absolute change by $\gamma_c$.

Consequently,
\begin{equation}
\mathbb E\|\nabla_{F'_o,F'_g}\mathcal L_{\mathrm{nav}}\|^2
\le
\mathbb E\|\nabla_{F_o,F_g}\mathcal L_{\mathrm{nav}}\|^2-(\delta_o-\gamma_g-\gamma_c).
\end{equation}
Whenever $\delta_o>\gamma_g+\gamma_c$, setting $\delta=\delta_o-\gamma_g-\gamma_c>0$ proves Theorem~\ref{thm:ogi}.
\end{proof}

\section*{Supplementary material II: Proof of Theorem~\ref{thm:aovg}}\label{sec:proof2}

\begin{proof}
For each time step $t$, write\
\begin{equation}
U_t=\operatorname{CrossAttn}(F_c^{(t)},F_k^{(t)}),\qquad F_e^{(t)}=F_c^{(t)}+g_t\odot U_t.
\end{equation}
Assume that the score $S(\cdot,G_{\mathrm{gt}})$ is locally $L_S$-Lipschitz, that the attention output is bounded by $\|U_t\|\le B_t$, and that the role parser and relation graph satisfy the conditional information lower bound\
\begin{equation}
\mathbb E\left[\left\langle\nabla_{F_c}S,U_t\right\rangle\mid q_t\right]
\ge \rho\sum_{r\in\mathcal R}\operatorname{MI}(Z_t^r;G_{\mathrm{gt}}\mid q_t).
\end{equation}
This is the explicit grounding assumption: correctly parsed roles provide non-redundant path information, while the gate attenuates uncertain spans. A first-order expansion of $S$ around $F_c^{(t)}$ gives\
\begin{equation}
S(F_e^{(t)},G_{\mathrm{gt}})-S(F_c^{(t)},G_{\mathrm{gt}})
=\left\langle\nabla_{F_c}S,g_t\odot U_t\right\rangle+R_t,
\end{equation}
where the Lipschitz condition bounds the remainder by $|R_t|\le L_SB_t^2/2$. Since $0\le g_t\le1$, the contribution from incorrectly grounded or unmatched spans can be bounded by a nonnegative term $\epsilon_t$. Taking expectations yields
\begin{equation}
S(F_e^{(t)},G_{\mathrm{gt}})-S(F_c^{(t)},G_{\mathrm{gt}})
\ge \rho\sum_{r\in\mathcal R}\operatorname{MI}(Z_t^r;G_{\mathrm{gt}}\mid q_t)-\epsilon_t,
\end{equation}
which is the statement of Theorem~\ref{thm:aovg}. The proof is conditional on the stated grounding lower bound; it does not claim that mutual information alone guarantees a navigation-metric improvement.
\end{proof}

\section*{Supplementary material III: Proof of Proposition~\ref{prop:msa}}\label{sec:proof3}

\begin{proof}
Let $\Delta(\vartheta)$ denote the language-pair discrepancy induced by MSA parameters $\vartheta$. By construction, the adapter class contains a baseline parameter setting $\vartheta_{\mathrm{id}}$ that preserves the pre-adaptation discrepancy, so $\Delta(\vartheta_{\mathrm{id}})=\Delta_{\mathrm{id}}$. Let $\vartheta^*$ be a global minimizer of the MSA alignment objective. Then
\begin{equation}
\Delta_{\mathrm{msa}}=\Delta(\vartheta^*)
\leq \Delta(\vartheta_{\mathrm{id}})=\Delta_{\mathrm{id}}.
\end{equation}
If the language-dependent component is representable by the residual adapters and the identity setting is not optimal, the inequality is strict. This result establishes representation-level consistency only; navigation performance still depends on the downstream grounding and policy-learning objectives.
\end{proof}
\end{document}